\documentclass[pdflatex,sn-mathphys-num, iicol]{sn-jnl}% Math and Physical Sciences Numbered Reference Style 

\usepackage{graphicx}
\usepackage{booktabs}
\usepackage{natbib}
\usepackage{orcidlink}
\usepackage{multirow}%
\usepackage{amsmath,amssymb,amsfonts}%
\usepackage{amsthm}%
\usepackage{mathrsfs}%
\usepackage[title]{appendix}%
\usepackage{xcolor}%
\usepackage{textcomp}%
\usepackage{manyfoot}%
\usepackage{booktabs}%
\usepackage{algorithm}%
\usepackage{algorithmicx}%
\usepackage{algpseudocode}%
\usepackage{listings}%
\usepackage{caption}
\usepackage{framed}
\usepackage{changepage}
\definecolor{citecolor}{HTML}{2980b9}
\definecolor{linkcolor}{HTML}{c0392b}
\newenvironment{formal}{%
  \MakeFramed{\advance\hsize-\width\FrameRestore}%
  \noindent\hspace{-4.55pt}%
  \begin{adjustwidth}{5pt}{5pt}%
  \vspace{5pt}
}
{%
  \vspace{5pt}\end{adjustwidth}\endMakeFramed%
}

\usepackage{orcidlink}
\usepackage{url}
\usepackage{graphicx}
\usepackage{amsmath}
\usepackage{wrapfig}
\usepackage{amssymb}
\usepackage{booktabs}
\usepackage{adjustbox}
\usepackage{chngcntr}
\usepackage{multirow}
\usepackage{makecell}

\usepackage[accsupp]{axessibility}
\usepackage{xcolor}
\definecolor{citecolor}{HTML}{2980b9}
\definecolor{linkcolor}{HTML}{c0392b}
\definecolor{dorange}{HTML}{ff6103}
\makeatletter
  \newcommand\figcaption{\def\@captype{figure}\caption}
  \newcommand\tabcaption{\def\@captype{table}\caption}
\makeatother
\usepackage{color, colortbl}
\usepackage{float}
\usepackage[capitalize]{cleveref}
\crefname{section}{Sec.}{Secs.}
\Crefname{section}{Section}{Sections}
\Crefname{table}{Table}{Tables}
\crefname{table}{Tab.}{Tabs.}

\begin{document}
\author[1]{\fnm{Ziyu} \sur{Guo} \orcidlink{0000-0001-9606-9691}} \email{guoziyu86@gmail.com} \equalcont{These authors contributed equally to this work.}

\author*[1]{\fnm{Renrui} \sur{Zhang} \orcidlink{0000-0003-4503-5277}} \equalcont{These authors contributed equally to this work.} \email{1700012927zrr@gmail.com}

\author[3]{\fnm{Chengzhuo} \sur{Tong}} \equalcont{These authors contributed equally to this work.}

\author[2]{\fnm{Zhizheng} \sur{Zhao}} \equalcont{These authors contributed equally to this work.}

\author[4]{\fnm{Rui} \sur{Huang}}

\author[3]{\fnm{Haoquan} \sur{Zhang}}

\author[1]{\fnm{Manyuan} \sur{Zhang}}

\author[2]{\fnm{Jiaming} \sur{Liu}}

\author[2]{\fnm{Shanghang} \sur{Zhang}}

\author[3]{\fnm{Peng} \sur{Gao}}

\author[1]{\fnm{Hongsheng} \sur{Li} \orcidlink{0000-0002-2664-7975}} %\email{hsli@ee.cuhk.edu.hk}

\author[1]{\fnm{Pheng-Ann} \sur{Heng} \orcidlink{0000-0003-3055-5034}} %\email{pheng@cse.cuhk.edu.hk}

% Affiliations
\affil[1]{\orgname{The Chinese University of Hong Kong}, \orgaddress{\postcode{999077}, \country{Hong Kong SAR}}}

\affil[2]{\orgname{Peking University}, \orgaddress{\city{Beijing}, \country{China}}}

\affil[3]{\orgname{Shanghai AI Laboratory}, \orgaddress{\city{Shanghai}, \postcode{200232}, \country{China}}}

\affil[4]{\orgname{University of Electronic Science and Technology of China}, \orgaddress{\city{Chengdu}, \country{China}}}

\title[Article Title]{\centering\textit{Can We Generate Images with CoT?}\vspace{0.07cm}\\Let's Verify and Reinforce Image Generation Step by Step}

\abstract{Chain-of-Thought (CoT) reasoning has been extensively explored in large models to tackle complex understanding tasks. However, it still remains an open question whether such strategies can be applied to verifying and reinforcing image generation scenarios. In this paper, we provide \textit{the first} comprehensive investigation of the potential of CoT reasoning to enhance autoregressive image generation. We focus on three techniques: scaling test-time computation for verification, aligning model preferences with Direct Preference Optimization (DPO), and integrating these techniques for complementary effects. Our results demonstrate that these approaches can be effectively adapted and combined to significantly improve image generation performance.
Furthermore, given the pivotal role of reward models in our findings, we propose the \textbf{P}otential \textbf{A}ssessment \textbf{R}eward \textbf{M}odel (\textbf{PARM}) and \textbf{PARM++}, specialized for autoregressive image generation. PARM adaptively assesses each generation step through a potential assessment approach, merging the strengths of existing reward models, and PARM++ further introduces a reflection mechanism to self-correct the generated unsatisfactory image, which is the first to incorporate reflection in autoregressive image generation. Using our investigated reasoning strategies, we enhance a baseline model, Show-o, to achieve superior results, with a significant +24\% improvement on the GenEval benchmark, surpassing Stable Diffusion 3 by +15\%.
We hope our study provides unique insights and paves a new path for integrating CoT reasoning with autoregressive image generation. Code and models are released at \url{https://github.com/ZiyuGuo99/Image-Generation-CoT}.}

\twocolumn
\maketitle
\twocolumn[
{%
\renewcommand\twocolumn[1][]{#1}
\begin{center}
\centering
\maketitle
\begin{minipage}[t]{\linewidth}
\centering
\includegraphics[width=\textwidth]{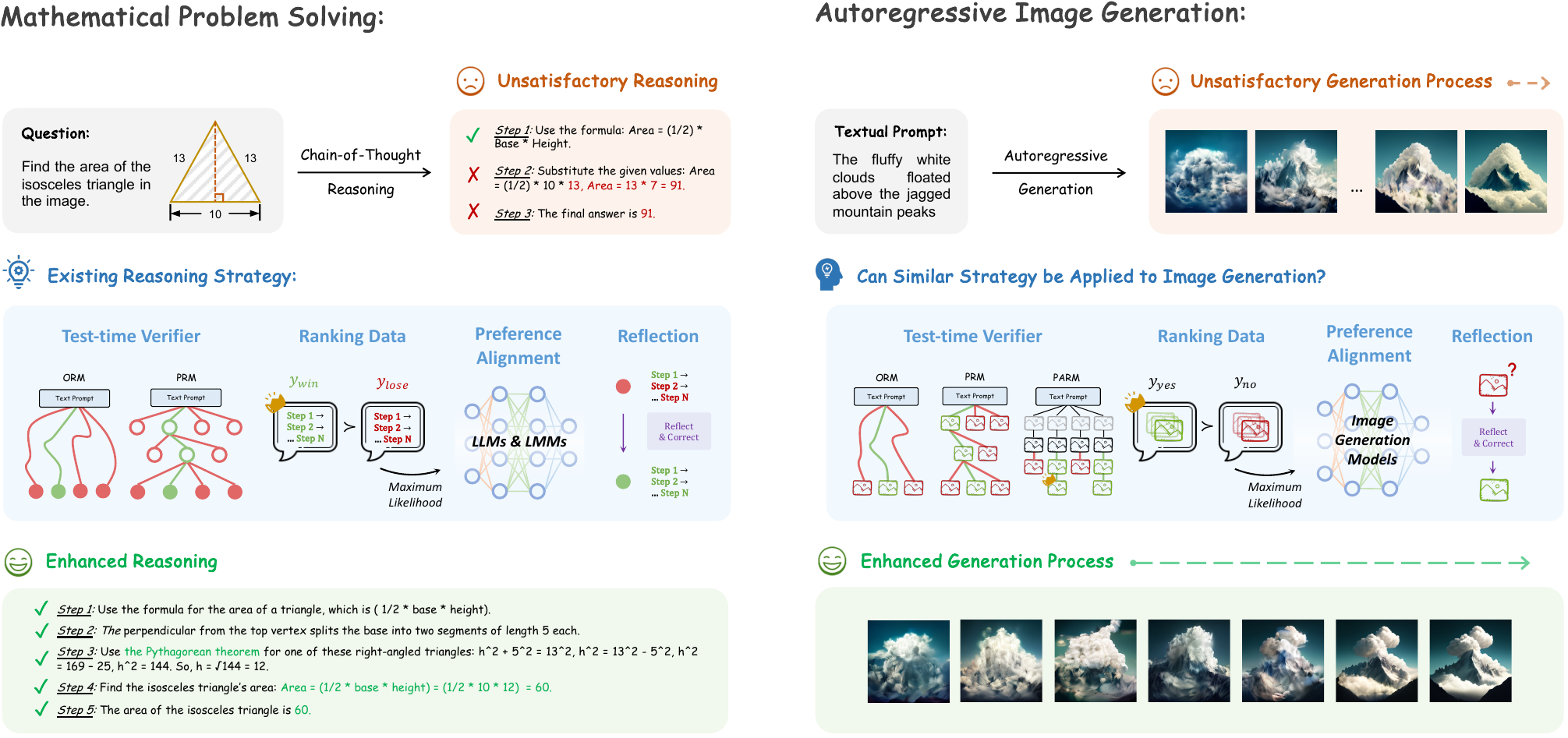}
\vspace{1pt}
{\captionsetup{hypcap=false}  
\captionof{figure}{{\textbf{\textit{Can We Verify and Reinforce Image Generation with Chain-of-Thought (CoT) Reasoning Strategies?}} Given the success of mathematical CoT reasoning in LLMs~\cite{lightman2023let,wei2022chain} and LMMs~\cite{zhang2024mathverse,zhang2024mavis} (Left), we provide \textit{the first} investigation to comprehensively explore the potential of applying current reasoning techniques to autoregressive image generation (Right), including test-time verification, preference alignment and reflection, with two newly proposed specialized reward models, termed \textbf{P}otential \textbf{A}ssessment \textbf{R}eward \textbf{M}odel (\textbf{PARM}) and PARM++.}}
\label{intro_v4}}
\end{minipage}
\end{center}
\vspace{0.8cm}
}]

\section{Introduction}
\label{sec:intro}
Large Language Models (LLMs)~\cite{touvron2023llama,touvron2023llama2,brown2020language} and Large Multi-modal Models (LMMs)~\cite{openai2023gpt4v,zhang2024llamaadapter,gao2023llamaadapterv2} have gained remarkable achievements across language~\cite{2023opencompass,OpenAI2023ChatGPT}, 2D image~\cite{fu2023mme,liu2023mmbench}, temporal video~\cite{fu2024video,han2023imagebind}, and 3D point cloud~\cite{guo2023point,guo2024sam2point}.
Building upon general understanding skills, recent efforts have been made toward enhancing LLMs and LMMs with complex Chain-of-Thought (CoT) reasoning capabilities~\cite{wang2022self,wei2022chain,zhang2023multimodal,jiang2024mmsearch}, e.g., OpenAI o1~\cite{gpt4o1}, contributing to superior performance in mathematics~\cite{zhang2024mathverse,Lu2023MathVistaEM}, science~\cite{sciverse,saikh2022scienceqa}, and coding~\cite{zhu2024deepseek,guo2024deepseek}.

Despite the success in multi-modal understanding, it remains under-explored \textit{whether multi-step reasoning strategies can be effectively applied to image generation}. Considering the discrepancy between two tasks, we observe that, autoregressive image generation~\cite{chang2022maskgit,xie2024show,sun2024beats,wang2024emu3} shares a similar output manner to the nature of LLMs and LMMs. Specifically, they both quantize the target data (language and image) into discrete tokens, and iteratively predict partial content conditioned on previously generated tokens. 

As illustrated in Figure~\ref{intro_v4}, LMMs leverage CoT to break down complex mathematical problems into manageable steps, which enables scaling test-time computation with reward models~\cite{snell2024scalingllmtesttimecompute,lightman2023let,ma2023let,wang2024math}  and reinforcement learning for preference alignment~\cite{zhang2024mavis,lu2024step,havrilla2024teaching,lai2024step}. Likewise, autoregressive image generation through step-by-step decoding can produce intermediate images, potentially allowing for similar verification and reinforcement techniques.
This raises the question: \textit{Can we verify and reinforce image generation step-by-step with strategies revealed by OpenAI o1?}

To this end, we conduct a systematic investigation into the potential of CoT reasoning for autoregressive image generation. We adopt Show-o~\cite{xie2024show}, a latest discrete generative model, as our baseline, and evaluate on a challenging text-to-image generation benchmark: GenEval~\cite{ghosh2024geneval}. Specifically, we focus on examining two key perspectives: \textit{1) Scaling test-time computation with Outcome/Process Reward Model (ORM/PRM) as verifiers}; and \textit{2) Reinforced preference alignment via Direct Preference Optimization (DPO)}. The specifics are as follows:

\begin{figure*}[t!]
\includegraphics[width=\linewidth]{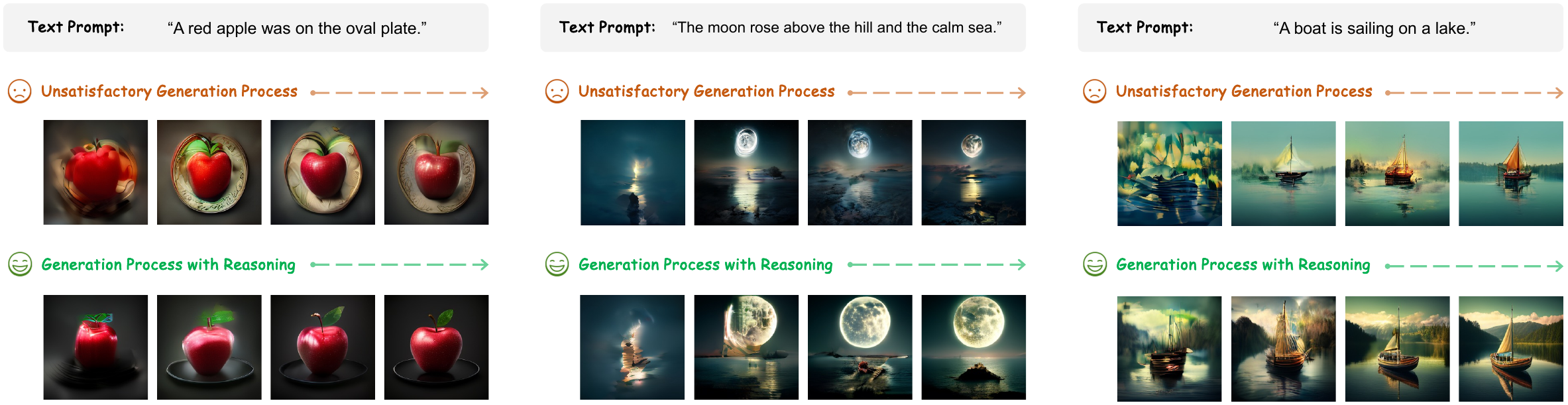}
\vspace{1pt}
\caption{\textbf{Autoregressive Image Generation without (Top) and with (Bottom) Our Reasoning Strategies.} We adopt Show-o~\cite{xie2024show} as the baseline model that produces unsatisfactory text-to-image generation. After using our investigated reasoning strategies (integrating PARM with iterative DPO for both reward model guidance and test-time verification), the generation process is effectively enhanced.}
\label{intro_vis}
\end{figure*}

\begin{itemize}
    \item \textbf{ORM \textit{vs} PRM as Test-time Verifiers.}
    As the top-1 result may not always be reliable, reward models are employed to score sampled candidates and perform outcome selection, where ORM is instance-level and PRM is process-level. In our settings, the score assesses whether each candidate image is inherently reasonable and aligns with the given textual prompt. We prompt LLaVA-OneVision (7B)~\cite{li2024llava} as a zero-shot reward model, and then curate text-to-image ranking data for reward fine-tuning. We apply a best-of-$N$ selection approach in the comparison of zero-shot and fine-tuned reward models.\vspace{0.1cm}
    
    \textit{\textbf{Observation:} ORM demonstrates significant boost, while PRM offers minimal benefit.}

\vspace{0.8em}
    \item \textbf{Test-time Verifiers \textit{vs} Preference Alignment.}
    Exploring the trade-off between inference-time and post-training offers valuable insights into the model's attainable performance. Preference alignment are adopted to elicit the implicit reasoning capabilities from their widely learned knowledge.
    In this study, we construct ranking preference data and apply DPO alignment with iterative training~\cite{pang2024iterative} to optimize the generation decoding process, comparing its effectiveness to test-time verification.\vspace{0.1cm}

    \textit{\textbf{Observation:} DPO alignment with iterative training attains stronger results to the fine-tuned ORM verifier.}

\vspace{0.8em}
    \item \textbf{Preference Alignment \textit{plus} Test-time Verifiers.}
    After investigating the individual impact, we integrate the two techniques to highlight their complementary potential in autoregressive image generation. We consider three approaches: \textit{1) DPO with reward model guidance}, i.e., integrating DPO’s policy with reward models' objectives for alignment; \textit{2) Verification after DPO alignment}, i.e., applying reward models for best-of-$N$ selection on DPO-aligned models; and \textit{3) Verification after DPO with reward model guidance}, which is a combination of \textit{1)} and \textit{2)}.\vspace{0.1cm}

    \textit{\textbf{Observation:} All integration methods lead to greater improvements, indicating complementary characteristics.}

\end{itemize}

Through our experiments, we demonstrate the promsing potential of applying CoT reasoning strategies to image generation scenarios, uncovering their adaptation methods and compatibility.
Furthermore, we identify significant room for improvement in reward models tailored for autoregressive image generation. For ORM, the global-level assessments are unable to capture the nuanced step-wise information, leading to inaccurate reward judgments. For PRM, the early-stage images tend to appear blurry, while later-stage images across different paths often converge to visually similar outputs, severely limiting its discrimination ability.

To alleviate these issues, we propose a specialized reward model for autoregressive image generation, termed
\textbf{P}otential \textbf{A}ssessment \textbf{R}eward \textbf{M}odel (\textbf{PARM}). PARM adaptively verifies the generation process step by step with three delicately designed tasks: \textit{1) Judge which step is clear and convincing enough to be evaluated}, given that most early-stage images are typically blurry; \textit{2) Assess whether the current step has the potential to yield a high-quality final image}, since later-stage images generally do not change too much; and \textit{3) Score the remaining final paths for selecting the best one}, similar to an ORM. In this way, PARM can adaptively conduct assessment at appropriate steps (overcoming PRM’s scoring challenges), while effectively capturing fine-grained step-by-step cues (complementing the coarse ORM). Our experiments showcase that PARM significantly outperforms both ORM and PRM, which finally improves the baseline model by +24\% on GenEval, as visualized in Figure~\ref{intro_vis}, surpassing the advanced Stable Diffusion 3~\cite{esser2024scalingrectifiedflowtransformers} by +15\%. {To further harness CoT-style reasoning, we introduce a reflection mechanism in PARM++, enabling the model to self-evaluate and refine its outputs in a multi-step, introspective manner.}

In addition, we further introduce an enhanced variant, \textbf{PARM++}, equipping PARM with a reflection mechanism that enables autoregressive models to refine its previously generated images. 
% Building upon the three tasks of PARM, PARM++ incorporates a reflection-inspired CoT evaluation step, where the model assesses the alignment of the generated image with the text prompt and reflects on potential issues—mimicking a self-questioning and correction process akin to human-like CoT reasoning.
Building upon the three tasks of PARM, PARM++ incorporates a reflection evaluation step to assess whether the generated image aligns with the input text prompt. If any misalignment is detected, PARM++ identifies the issue with detailed descriptions, e.g., discrepancies in visual concepts, and prompts the generative model to iteratively self-correct the image until it passes the reflection evaluation. By refining the text-to-image generation process, PARM++ achieves a +4\% improvement over PARM on the baseline model on GenEval.

Our contributions are summarized as follows:
\begin{itemize}
    \item We present \textit{the first} comprehensive empirical study of applying CoT reasoning strategies to autoregressive image generation domains, providing unique insights into the future advancement of this field.

    \vspace{0.5em}
    \item We investigate specific adaption methods of techniques, including test-time verification, preference alignment, ranking data curation and reflection, to autoregressive image generation, indicating their performance and complementarity. 

    \vspace{0.5em}
    \item We further introduce PARM and PARM++, two new reward models tailored for image generation scenarios, which adaptively perform step-wise potential assessment and reflection evaluation for self-correction, significantly enhancing text-to-image generation quality.
    
\end{itemize}

\section{Related Work}

\paragraph{Scaling Test-time Computation.}
Humans often dedicate significant time and effort to solve complex problems. Inspired by this, many efforts have focused on scaling test-time computation for Large Language Models (LLMs) to tackle reasoning tasks such as mathematical problem-solving~\cite{wang2024math, zhang2024rest, xin2024deepseek, sun2024beats}, code synthesis~\cite{delorenzo2024make, ni2023lever, zhu2024deepseek}, and workflow generation~\cite{zhang2024aflow, gal2024comfygen, xue2024genagent}.
One line of research adapts the input space to leverage Chain-of-Thought (CoT) capabilities, using approaches like in-context CoT examples~\cite{wei2022chain} or zero-shot CoT prompts~\cite{kojima2022large}. Another branch modifies or integrates reasoning paths within the output space, utilizing strategies such as self-consistency~\cite{wang2024chain}, CoT decoding~\cite{wang2024chain}, and verifier-based selection~\cite{cobbe2021training, lightman2023let, snell2024scalingllmtesttimecompute}. Among these, test-time verifiers have demonstrated generality and robustness in enhancing reasoning performance.
For example, early work~\cite{cobbe2021training} trains an Outcome Reward Model (ORM) to evaluate final outputs and select the best-of-$N$ candidates for optimal results. Later, Lightman et al.~\cite{lightman2023let, ma2023let} adopt the Process Reward Model (PRM) to evaluate intermediate reasoning steps, achieving greater effectiveness. Snell et al.~\cite{snell2024scalingllmtesttimecompute} further highlights that scaling test-time computation is often more impactful than scaling model parameters during training.
Recently, OpenAI o1~\cite{openai2024learning} has demonstrated exceptional reasoning capabilities across a variety of complex and challenging scenarios, underscoring the potential of this approach. Building on these advancements in understanding tasks, we conduct a comprehensive investigation into whether verifier-based strategies can also enhance image generation tasks, and propose a new Potential Assessment Reward Model (PARM), specifically designed for this domain.

\paragraph{Reinforced Preference Alignment.}
After robust pre-training and fine-tuning, LLMs often acquire substantial knowledge. However, a post-training alignment stage is typically required to align their output preferences to meet specific targets, such as human feedback~\cite{christiano2017deep, bai2022constitutional, kupcsik2018learning} or Chain-of-Thought (CoT) reasoning~\cite{wang2024math, lai2024step, lu2024step}.
Traditional approaches~\cite{pacchiano2021dueling, jain2013learning, busa2014preference, zhao2023slic} often leverage reinforcement learning (RL) to address this challenge. These methods usually involve two steps: first, optimizing a neural-network-based reward function within a preference model (e.g., the Bradley-Terry model~\cite{bradley1952rank}), and then fine-tuning the target LLM to maximize this reward using techniques like proximal policy optimization (PPO)~\cite{schulman2017proximal}. However, RL-based methods often encounter issues related to complexity and instability.
To overcome these challenges, Rafailov et al. introduced Direct Preference Optimization (DPO)~\cite{rafailov2024direct}, which parameterizes the reward model to enable the derivation of the optimal policy through a closed-form solution. This approach has been effectively applied to enhance CoT capabilities in mathematical reasoning~\cite{luo2023wizardmath, wang2024math} and code generation~\cite{xu2024dpo, miao2024aligning, gee2024code}. Further advancements have extended DPO with step-wise preference data~\cite{lai2024step, lu2024step} for more granular supervision and multi-modality learning~\cite{zhang2024mavis, zhang2024improve} to support visual reasoning.
In this study, we apply DPO-based preference alignment to autoregressive image generation, demonstrating its effectiveness in improving image quality during step-by-step decoding.

\paragraph{Autoregressive Image Generation.}
The transformer architectures with autoregressive output schemes~\cite{llama3modelcard,li2024llava,touvron2023llama2,openai2023gpt4v,openai2024gpt4o,OpenAI2023GPT4TR} have demonstrated a remarkably successful modeling approach in language and multi-modality.
Motivated by such progress, a series of work, e.g., DALL-E~\cite{ramesh2021zero}, LlamaGen~\cite{sun2024autoregressive}, and Chameleon~\cite{chameleonteam2024chameleonmixedmodalearlyfusionfoundation}, utilizes such autoregressive modeling with casual attention to learn the dependency within image pixels for image generation tasks, rather than popular diffusion models~\cite{chen2024pixartalpha,ramesh2022hierarchicaltextconditionalimagegeneration,esser2024scalingrectifiedflowtransformers,podell2023sdxlimprovinglatentdiffusion,zhou2024transfusion,jiang2024comat}.
However, such raster-order autoregression suffers from severe time consumption and performance constraints when synthesizing high-resolution and high-fidelity images, attributed to the growing number of discrete tokens compressed by VQ-VQE~\cite{van2017neural,razavigenerating,gu2022vector,esser2021taming}.
To address the challenges, MaskGiT~\cite{chang2022maskgit} proposes to learn a bidirectional autoregressive transformer with a parallel iterative decoding strategy, benefiting both the generation performance and efficiency.
Recently, this approach has been effectively extended, primarily focusing on two aspects: the unification of visual understanding and generation (Show-o~\cite{xie2024show}) and its integration with diffusion techniques (MAR~\cite{li2024autoregressive}).
Considering that such generation paradigm is quite similar to that of LLMs, representing data with discrete tokens and predicting iteratively conditioned on previous tokens, we explore the potential of applying CoT reasoning techniques within LLMs to autogressive image generation. Through our thorough investigation, we demonstrate its promising effectiveness for enhanced image generation capabilities.

\section{Our Investigation}
\label{sec:Methodology}

%-------------------------------------------------------------------------
Chain-of-Thought (CoT) reasoning has been widely exploited to solve complex problems for language and multi-modal understanding. In this study, we conduct a systematic investigation aiming to find out, whether we can verify and reinforce image generation step-by-step.

% In Sec.~\ref{}, we first illustrate our problem definition and experimental settings. Then, in

\subsection{Overview}

\begin{figure}[t]
  \centering
  \includegraphics[width=0.9\linewidth]{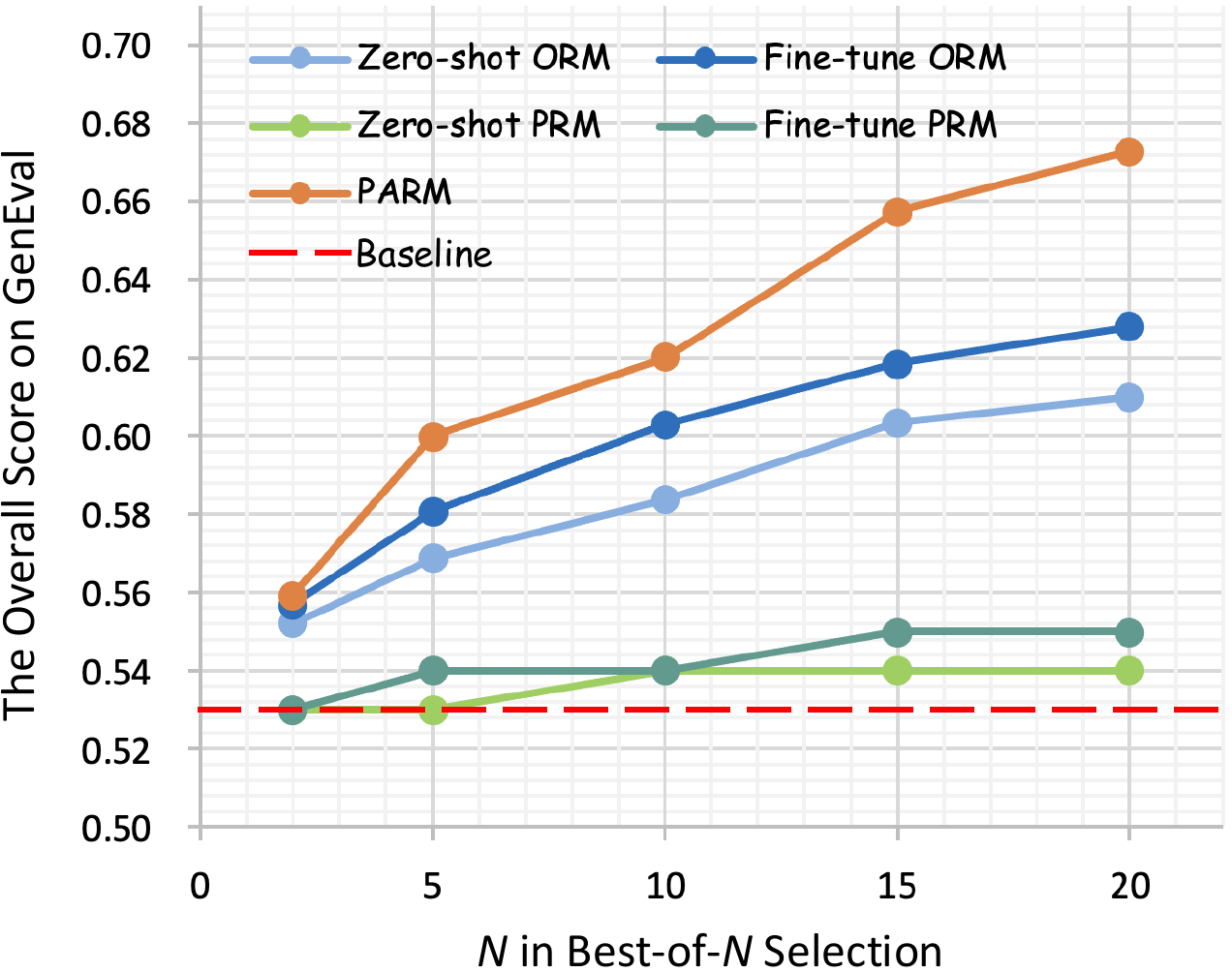}
    \vspace{0.2cm}
  \caption{\textbf{Comparison of Reward Models as Test-time Verifiers.} We adopt Show-o~\cite{xie2024show} as the `Baseline' and evaluate Best-of-$N$ selection on the GenEval~\cite{ghosh2024geneval} benchmark.}
  \label{bon}
% \vspace{-0.05cm}
\end{figure}

\begin{figure*}[t!]
\centering
\includegraphics[width=\textwidth]{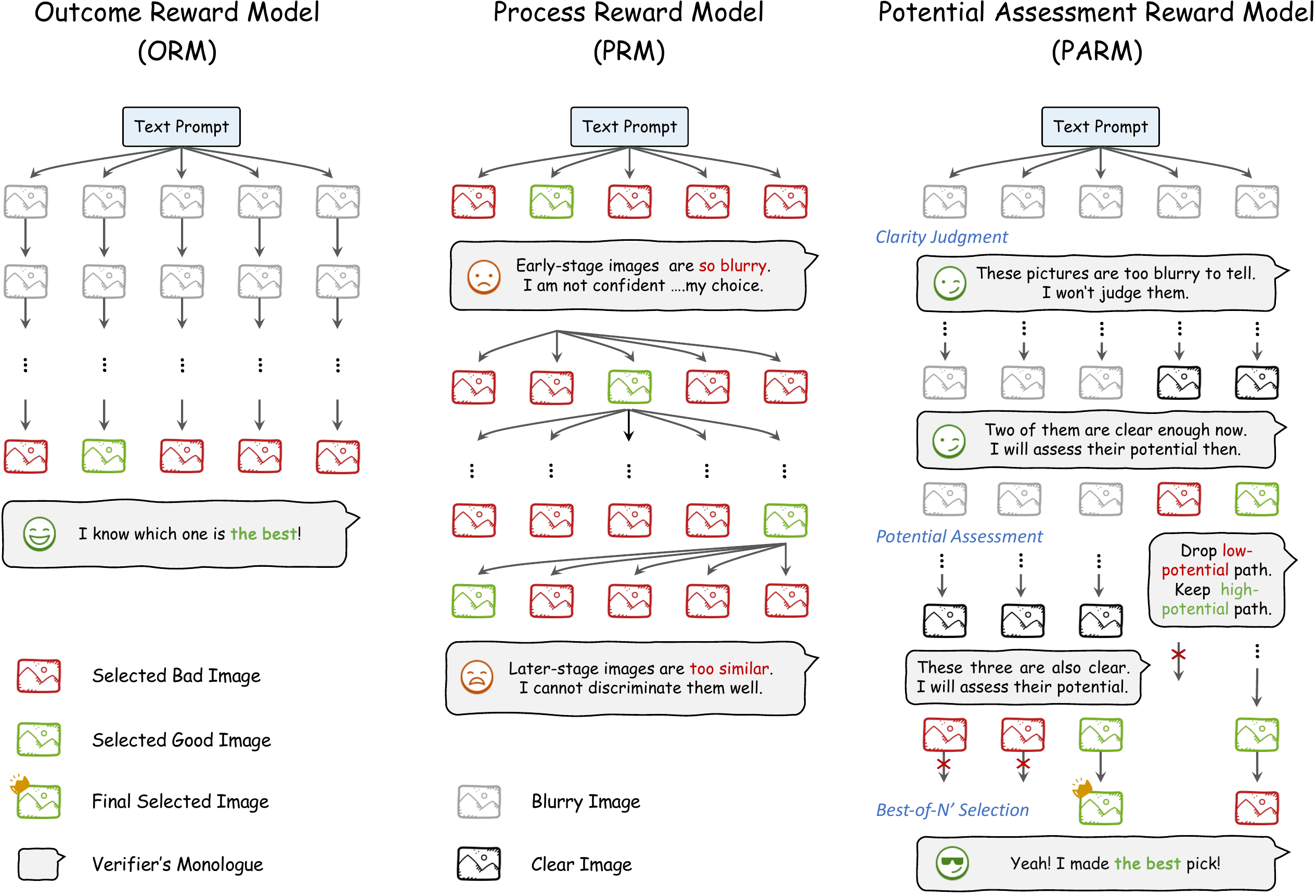}
\vspace{0.2cm}
   \caption{\textbf{Investigation of Reward Models in Autoregressive Image Generation.} For test-time verification, we implement Outcome Reward Model (ORM) and Process Reward Model (PRM), and introduce a new Potential Assessment Reward Model (PARM) customized for image generation scenarios, which progressively performs three tasks (highlighted in blue) to enhance the reasoning of generation.}
\label{rm_v1}
\end{figure*}

\paragraph{Task Formulation.} 
To enable the applicability of current reasoning techniques, we focus on the autoregressive image generation task, demonstrated by models such as MaskGiT~\cite{chang2022maskgit} and LlamaGen~\cite{sun2024beats}. This task employs a data representation and output paradigm akin to those used in LLMs and LMMs, while achieving comparable performance to continuous diffusion models~\cite{rombach2022high,ramesh2022hierarchicaltextconditionalimagegeneration,podell2023sdxlimprovinglatentdiffusion}. Specifically, it leverages quantized autoencoders~\cite{esser2021taming} to transform images into discrete tokens, allowing for the cross-entropy loss of Direct Preference Optimization (DPO)~\cite{rafailov2024direct} in post-training. Additionally, it iteratively predicts one or more tokens at each step, conditioned on prior outputs, thereby creating reasoning paths suitable for step-wise verification with reward models.

\paragraph{Experimental Settings.}
We select Show-o~\cite{xie2024show} as our baseline model for investigation, a latest autoregressive image generation model with advanced capabilities. To comprehensively evaluate different strategies, we assess the text-to-image generation performance on a rigorous benchmark: GenEval~\cite{ghosh2024geneval}. This scenario challenges the model to produce images with not only high visual quality and image-text alignment, but also accurate object attribute and co-occurrence. 
In the subsequent sections, we explore three strategies to improve the step-by-step decoding of image generation: test-time verification (Sec.~\ref{s3.2}), preference alignment (Sec.~\ref{s3.3}), and their combination (Sec.~\ref{s3.4}).

\subsection{ORM \textit{\textbf{vs}} PRM as Test-time Verifiers}
\label{s3.2}

Scaling test-time computation~\cite{snell2024scalingllmtesttimecompute,lightman2023let,ma2023let,wang2024math} to enhance reasoning capabilities has emerged as an effective alternative to scaling training costs. Current approaches often employ reward models as test-time verifiers within CoT reasoning paths, typically using two main categories: Outcome Reward Model (ORM) and Process Reward Model (PRM). Drawing inspiration from these methods, we respectively implement and evaluate them within the context of autoregressive image generation, as illustrated in Figure~\ref{rm_v1}.

\paragraph{ORM.} Based on multiple complete reasoning outputs, ORM assigns each candidate a reword score and select the most confident one using a best-of-$N$ strategy. 
In our study, we adopt ORM solely to evaluate the generated image at the final step, rather than the entire CoT process in mathematical reasoning tasks.
Specifically, we begin with a zero-shot ORM, followed by curating a text-to-image ranking dataset to fine-tune the ORM for enhancement, as outlined below:

\begin{itemize}
    \item \textbf{\textit{Zero-shot ORM:}} We leverage a pre-trained LLaVA-OneVision (7B)~\cite{li2024llava}, an LMM with superior generalization, as our zero-shot ORM. We input the text prompt along with the generated image into LLaVA-OneVision, and devise a prompt template to activate its visual understanding capabilities for text-to-image evaluation, which we observe performs well in most cases, as below:
\begin{formal}
    \textbf{Prompt:} \textit{``\textbf{ \{image\}} This image is generated by a prompt: \textbf{\{prompt\}}. Does this image accurately represent the prompt? Please answer yes or no without explanation.''}
\end{formal}
\noindent The `\textbf{\textit{\{image\}}}' and `\textbf{\textit{\{prompt\}}}' denote and generated image from Show-o~\cite{xie2024show} and the input textual prompt. 
This model assesses the quality of candidate images, providing binary responses, `yes' (good quality) or `no' (low quality). The candidate image with the highest probability of `yes' is then selected as the final output.
        \vspace{0.5em}

    \item \textbf{\textit{ORM Ranking Data Curation:}} To enhance the accuracy of outcome rewards, we curate a dataset of 288K text-to-image ranking examples for fine-tuning ORM. First, we prompt GPT-4~\cite{OpenAI2023GPT4TR} to generate a list of 200 countable daily object names with specific colors. Using these objects, we apply the six object-centered prompt templates from GenEval, constructing a diverse set of 13K text prompts. We perform a strict filtering to ensure that these prompts do not overlap with the GenEval test samples. Then, using our baseline model, Show-o, we synthesize around 50 images per prompt at a high temperature,  %After that, we label each image with a binary annotation of `yes' or `no' using the evaluation metric in GenEval.
    % We adopt the same prompt in the instruction as the zero-shot ORM, 
    and label `yes' or `no' in the response to denote the positive or negative instance, as showcased below:
    \begin{formal}
        \textbf{Instruction:} \textit{``\textbf{ \{image\}} This image is generated by a prompt: \textbf{\{prompt\}}. Does this image accurately represent the prompt? Please answer yes or no without explanation.''}\\
        \\
        \textbf{Response:} \textit{``Yes''} or \textit{``No''}
    \end{formal}

    \vspace{0.5em}
    
    \item \textbf{\textit{Fine-tuned ORM:}} Using the curated ranking dataset, we fine-tune LLaVA-OneVision to enhance its capability for assessing image quality and cross-modal alignment. The training data format is consistent with the prompt template used in the zero-shot ORM, incorporated with our constructed 288K text prompts and associated images. The model is fine-tuned for one epoch, using a batch size of $8$ and a learning rate of $1e^{-5}$. This fine-tuning process enables the ORM to capture more intricate aspects of object composition and nuanced visual-text relationships, resulting in more reliable scoring.
    
\end{itemize}

\paragraph{PRM.}
Different from ORM that evaluates only the final output, we utilize PRM to provide a reward score to each candidate with different steps throughout the generation process.
Similar to our previous investigation, we start from a zero-shot PRM, LLaVA-OneVision,
and then curate 10K step-wise text-to-image ranking data to obtain a fine-tuned PRM. 

%%%%%%%%%%%%%
\begin{itemize}
    \item \textbf{\textit{Zero-shot PRM:}} 
We also utilize the pre-trained LLaVA-OneVision (7B) as our zero-shot PRM, applying a similar prompt template used in ORM as:
\begin{formal}
    \textbf{Prompt:} \textit{``\textbf{ \{image\}} This is an intermediate image in the generation process by a prompt: \textbf{\{prompt\}}. Does this intermediate image accurately represent the prompt? Please answer yes or no without explanation.''}
\end{formal}
\noindent At each intermediate step in the generation process, the zero-shot PRM assesses each candidate image with a binary response, `yes' or `no'. We then adopt a step-level best-of-$N$ strategy, selecting the most confident candidate and following this path for subsequent decoding. By iteratively employing the PRM at each step, the generation process is guided step by step towards the final.

    \vspace{0.5em}

\item \textbf{\textit{PRM Ranking Data Curation:}}
We observe that the images generated at intermediate steps tend to appear very blurry, as only partial visual tokens in specific regions are decoded while others remain unresolved. Since LLaVA-OneVision is pre-trained only on natural images (similar to those generated at the final step), the zero-shot PRM has limited capability for precise step-wise evaluation. To address this issue, we curate a 300K step-wise text-to-image ranking dataset to fine-tune an improved PRM. We adopt the same prompt in the instruction as the zero-shot PRM, formulated as:
\begin{formal}
    \textbf{Instruction:} \textit{``\textbf{ \{image\}} This is an intermediate image in the generation process by a prompt: \textbf{\{prompt\}}. Does this intermediate image accurately represent the prompt? Please answer yes or no without explanation.''}\\
    \\
    \textbf{Response:} \textit{``Yes''} or \textit{``No''}
\end{formal}
\noindent First, we utilize the 13K unique text prompts from our ORM ranking dataset, generating 18 intermediate-step images per prompt using Show-o. Inspired by Math-Shepherd~\cite{wang2024math}, we employ an automated annotation approach to obtain accurate step-wise labels, eliminating the need for costly human labor or GPT assistance. For instance, to label the image at step $i$ ($1\leq i \leq 18$), we condition Show-o on that image and then produce four different paths for the remaining 18 - $i$ steps. By evaluating the final images from each of these paths, if any path receives a `yes' score, it indicates that step $i$ has a high potential to lead to a correct final output, and thus it is labeled as `yes'; otherwise, it is labeled as `no'. This automated approach allows us to efficiently obtain step-wise annotations for assessing the generation.

\begin{figure*}[t]
  \centering
  \includegraphics[width=0.9\linewidth]{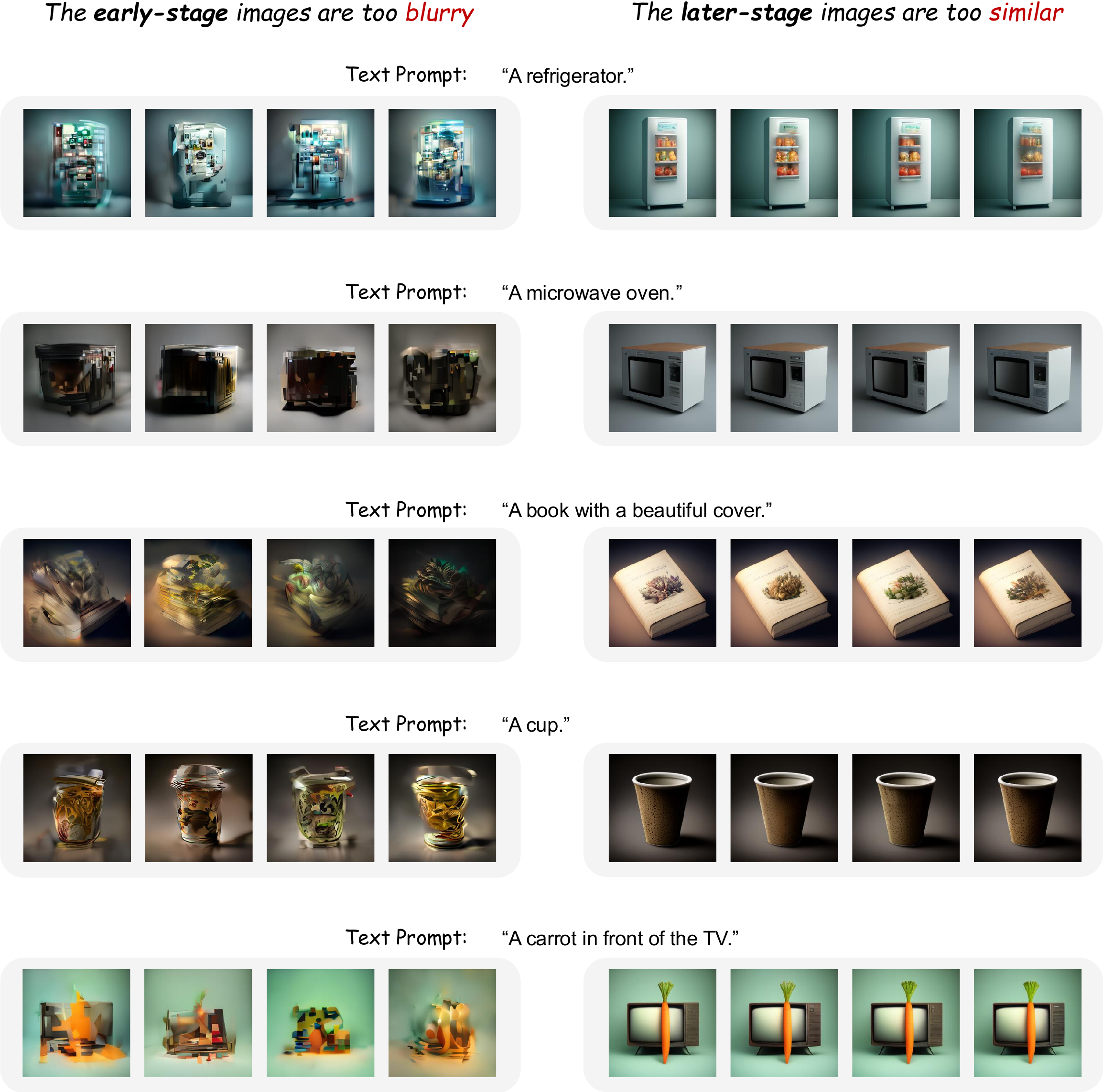}
  \vspace{0.2cm}
  \caption{\textbf{Visualization of Early-stage and Later-stage Images.} We visualize the generated images in the intermediate steps of Show-o~\cite{xie2024show}, where the early-stage images are too blurry to interpret, while the later-stage images are too similar to discriminate, posing great challenges for PRMs to evaluate.}
  \label{early-later}
\end{figure*}

    \vspace{0.5em}

\item \textbf{\textit{Fine-tuned PRM:}}
With the step-wise ranking data, the LLaVA-OneVision is fine-tuned to boost the visual comprehension of intermediate-step images. The data format and training configurations are the same as those used for fine-tuning the ORM. After training, the PRM becomes more capable of interpreting blurry images within the decoding process for more accurate step-by-step selection.
\end{itemize}

%%%%%%%%%%%%%

\paragraph{Experiments and Insights.} 
As showcased in the middle of Table~\ref{t1}, we compare the test-time verification results between ORM and PRM with a best-of-$20$ strategy. The observations are summarized below:
\begin{itemize}
    \item \textbf{\textit{Test-time verification can significantly boost generation performance.}} Compared to the baseline scores of 53\% on GenEval, the fine-tuned ORM as a test-time verifier achieves the highest gains of +10\%. This finding suggests that current autoregressive image generation models, similar to LLMs and LMMs, face challenges with inconsistent and unstable decoding paths. Consequently, a test-time verification strategy is essential to identify and follow the most reliable reasoning path.
    \vspace{0.5em}
    
    \item \textbf{\textit{ORM exhibits stronger enhancement capabilities than PRM.}}
    In contrast to ORM providing a clear benefit, PRM yields only marginal improvements, e.g., +2\% on GenEval after fine-tuning. This discrepancy arises from the unique characteristics of the autoregressive image generation task in two key ways:
    \textit{1)} Images at early steps are too blurry for PRM to effectively interpret their visual features and image-text alignment. 
    \textit{2)} Images at later steps tend to exhibit minimal differences, making it challenging for PRM to discriminate. 
    Whereas, ORM evaluates images at the final step, which provides sufficient visual and semantic information for accurate judgment.
    \vspace{0.5em}
    
    \item \textbf{\textit{Fine-tuning by ranking data enhances verification results and demonstrates improved scaling performance.}}
    As illustrated in Figure~\ref{bon}, both fine-tuned ORM and PRM outperform their zero-shot counterparts, achieving higher scores with larger $N$ values in the best-of-$N$ selection. Additionally, as $N$ increases, fine-tuned reward models show greater improvements, indicated by steeper curves, reflecting a better scaling response to test-time computation. This highlights the effectiveness of our curated ranking dataset in refining reward accuracy and benefiting scalability across a broader range of candidates.

\end{itemize}

\begin{table*}[t]
  \centering
  \small
  % \vspace{0.3cm}
  \caption{\textbf{Test-time Verifiers (ORM \textit{vs} PRM) \textit{vs} Preference Alignment.} We evaluate text-to-image generation on the GenEval~\cite{ghosh2024geneval} benchmark and adopt Show-o~\cite{xie2024show} as the autoregressive baseline model. `ORM/PRM' and `DPO' denote Outcome/Process Reward Model and Direct Preference Optimization~\cite{rafailov2024direct}, respectively. We adopt the best-of-$N$ selection for test-time verifiers, setting $N=20$, and highlight the better-performed variant of each reasoning strategy in green.}
  \vspace{0.1cm}
  \begin{tabular}{l|cc|cccccc|c}
    \toprule
    \makecell*[l]{Reasoning\\Strategy} &Method& Setting&\makecell*[c]{Single\\object} & \makecell*[c]{Two\\object} & \makecell*[c]{Counting} & \makecell*[c]{Colors} & \makecell*[c]{Position} & \makecell*[c]{Attribute\\binding} & \textbf{Overall} \\
    \midrule
    \rowcolor{gray!12}Baseline& - &- & 0.95 & 0.52 & 0.49 & 0.82 & 0.11 & 0.28 & 0.53 \\
    \cmidrule{1-10}
    \multirow{4}*{\makecell*[l]{Test-time\\Verifier}}&\multirow{2}*{ORM}  &Zero-Shot & 0.99 & 0.63 & 0.63 & 0.84 & 0.19 & 0.39 & 0.61 \\ 
&&Fine-tuned & 0.99 & 0.72 & 0.65 & 0.84 & 0.25 & 0.33 & \cellcolor{green!15}{0.63} \\
    \cmidrule{2-10}
    &\multirow{2}*{PRM} &Zero-Shot &0.98&0.51&0.54&0.82&0.11&0.23&0.53 \\ 
&&Fine-tuned &0.98&0.55&0.54&0.83&0.13&0.29&\cellcolor{green!15}{0.55} \\
    \cmidrule{1-10}
    Preference&\multirow{2}*{\makecell*[c]{DPO}} &- & 0.96 & 0.70 & 0.50 & 0.82 & 0.30 & 0.43 & 0.62 \\ 
    Alignment&&Iterative & 0.98 & 0.72 & 0.53 & 0.84 & 0.40 & 0.46 & \cellcolor{green!15}{0.65} \\ 
    \bottomrule
  \end{tabular}
  \label{t1}
  \vspace{0.2cm}
\end{table*}

\subsection{Test-time Verifiers \textit{\textbf{vs}} Preference Alignment}
\label{s3.3}

Post-training has been widely utilized in existing LLMs and LMMs to align model outputs with human preferences. 
Common techniques include reinforcement learning with reward models, e.g., Proximal Policy Optimization (PPO)~\cite{schulman2017proximal}, and its streamlined counterpart with classification objectives, e.g., Direct Preference Optimization (DPO)~\cite{rafailov2024direct}. 
Given that most autoregressive image generation models inherently function within a classification framework, we leverage the simplicity of DPO alignment to enhance the quality of generated images.

\paragraph{DPO Ranking Data Curation.}
To bypass reinforcement learning, DPO leverages an implicit rewarding mechanism by training on a ranking dataset of paired preferred and dispreferred responses, corresponding to well-generated and poor-quality images in our case. Fortunately, we have already constructed a substantial amount of ranking data for training ORM, annotated with `yes' and `no' labels to indicate the generation quality. Building on this, we utilize the 13K unique text prompts from the ORM training dataset and, for each prompt, randomly pair two generated images, one labeled `yes' and the other `no', yielding 10K paired data for DPO alignment. 

\paragraph{DPO for Autoregressive Image Generation.}
Since autoregressive image generation models are also trained using a cross-entropy loss, we can directly apply the maximum likelihood objective in DPO to our setting. In detail, the parameterized policy is initialized from Show-o and optimized during training, while the reference policy is also initialized from Show-o but kept frozen.
The objective encourages the model to assign a higher likelihood to preferred images over dispreferred images, aligning with the curated preference structure. The training is conducted over one epoch with a batch size of $10$ and a learning rate of $1e^{-5}$.

\paragraph{DPO with Iterative Training.}
Following the initial stage of DPO alignment, the model has learned to generate images that better align with the preferred responses. Inspired by iterative DPO~\cite{pang2404iterative}, we further refine this alignment by applying the newly aligned model to generate updated ranking data based on the text prompts in $\mathcal{D}$. We annotate these new images with `yes' or `no' labels using the same method in Sec.~\ref{s3.2}. For each prompt, we collect paired images labeled $y_{yes}$ and $y_{no}$, and exclude samples where all images receive the same label, resulting in a refined DPO ranking dataset of 7K samples.
By conducting another round of DPO, the model can be further improved by learning from more informative preference relations. We iterate the DPO training process once with the same training configurations.

\begin{table*}[t]
  \centering
  \small
  % \vspace{0.3cm}
  \caption{\textbf{Test-time Verifiers \textit{plus} Preference Alignment.} We evaluate text-to-image generation on the GenEval~\cite{ghosh2024geneval} benchmark and adopt Show-o~\cite{xie2024show} as the autoregressive baseline model. `Ft. ORM' and `It. DPO' denote the fine-tuned ORM and iterative DPO~\cite{rafailov2024direct}. We explore three combination approaches (`1\textit{st}, 2\textit{nd}, and 3\textit{rd} Integration') of reward models and preference alignment, comparing `individual' results. We adopt the best-of-$N$ selection for test-time verifiers, setting $N=20$, and highlight the best-performing integration in green.}
  \vspace{0.1cm}
  \begin{adjustbox}{width=\textwidth}
  \begin{tabular}{l|ccc|cccccc|c}
    \toprule
    \makecell*[l]{Reasoning\\Strategy} &\makecell*[c]{Test-time\\Verifier} & \makecell*[c]{Preference\\Alignment}& \makecell*[c]{Reward\\Guidance} &\makecell*[c]{Single\\object} & \makecell*[c]{Two\\object} & \makecell*[c]{Counting} & \makecell*[c]{Colors} & \makecell*[c]{Position} & \makecell*[c]{Attribute\\binding} & \textbf{Overall} \\
    \midrule
    \rowcolor{gray!12}{Baseline}&- &- &-& 0.95 & 0.52 & 0.49 & 0.82 & 0.11 & 0.28 & 0.53 \\
    \cmidrule{1-11}
    \multirow{2}*{Individual} & Ft. ORM& -&- & 0.99 & 0.72 & 0.65 & 0.84 & 0.25 & 0.33 & 0.63 \\
    &- & It. DPO &-& 0.98 & 0.72 & 0.53 & 0.84 & 0.40 & 0.46 & 0.65 \\
    \cmidrule{1-11}
    1\textit{st} Integration&- & It. DPO &Ft. ORM & 0.98 & 0.78 & 0.44 & 0.81 & 0.50 & 0.48 & 0.67 \\ 
    2\textit{nd} Integration&Ft. ORM & It. DPO &- & 0.98 & 0.80 & 0.62 & 0.83 & 0.59 & 0.54 &0.72 \\ 
    % \cmidrule{1-10}
    3\textit{rd} Integration &Ft. ORM & It. DPO &Ft. ORM &0.98 &0.84 &0.64 &0.85 &0.66 &0.52 &\cellcolor{green!15}0.75 \\ 
    \bottomrule
  \end{tabular}
  \end{adjustbox}
  \label{t2}
  \vspace{0.2cm}
\end{table*}

% \vspace{-0.2cm}
\paragraph{Experiments and Insights.}
In the bottom of Table~\ref{t1}, we present the evaluation results of DPO alignment and compare it with the performance with test-time verification. The observations are summarized below:
\begin{itemize}
    \item \textbf{\textit{DPO alignment can effectively reinforce the generation performance.}}
    On GenEval, initial DPO alignment improves the baseline model's performance by +9\%. With iterative training, these gains are further extended as +11\%, highlighting the effectiveness of a refined preference dataset in strengthening model alignment with desired outputs. This demonstrates that DPO alignment can serve as a powerful method for enhancing autoregressive generation models, especially in scenarios where explicit preference data is available to guide training.
    \vspace{0.5em}

    \item \textbf{\textit{Initial DPO matches test-time verification, while iterative DPO surpasses.}}  
    After the initial alignment, the model achieves performance comparable to that of the fine-tuned ORM, the top-performing variant for verification. However, with iterative alignment on refined ranking data, the model outperforms all test-time verifiers, i.e., +2\% over the fine-tuned ORM, indicating the potential of iterative DPO to progressively reinforce image generation capabilities through updated ranking data.

\end{itemize}

\subsection{DPO Alignment \textit{\textbf{plus}} Test-time Verifiers}
\label{s3.4}

The investigations above have demonstrated the individual effectiveness of test-time verification and DPO alignment. Next, we explore three approaches to integrate these two techniques to assess their complementary potential in image generation, leveraging both the adaptability of verifiers and the reinforcement of DPO.

 \paragraph{DPO with Reward Model Guidance.}
 As discussed in previous works~\cite{xu2024dpo}, DPO can struggle with out-of-distribution responses due to distribution shifts from the ranking dataset. A potential solution~\cite{llama3modelcard,qwen2} is to incorporate prompt-only datasets during post-training and leverage a reward model to provide online preference guidance. Following this approach, we adopt our fine-tuned ORM as the explicit reward model to offer more generalized preference feedback, and add the online objectives with the original DPO loss
We maintain the same training data and configurations as the initial DPO alignment stage.

 \paragraph{Verification after DPO Alignment.}
We observe that verification and DPO techniques may naturally complement each other in two key ways:
\textit{1)} They operate independently at different stages of implementation, i.e., post-training and test-time;
and \textit{2)} DPO refines the internal knowledge distribution within the model to enhance reasoning, while verification focuses on selecting the optimal reasoning path within this refined distribution. 
Therefore, we apply the fine-tuned ORM for best-of-$N$ selection directly on the model after DPO alignment.
 
 \paragraph{Verification after DPO with Reward Model Guidance.}
 In this approach, we combine the strengths of both DPO with reward model guidance and test-time verification. Our goal is to achieve optimal alignment, enhancing the model’s generalization capabilities during training, while also ensuring reliable image generation paths at inference time.
 
 \paragraph{Experiments and Insights.}
Table~\ref{t2} reports the text-to-image generation scores with different integration methods. The observations are summarized below:

\begin{itemize}
    \item \textbf{\textit{Verification and alignment perform expected complementary characteristics.}}
    Across all three approaches, verification and alignment complement each other effectively. For instance, the third integration method outperforms DPO alignment alone by +10\%, and surpasses the verification alone by +12\%. These results highlight the potential of combining verification and alignment techniques in future autoregressive image generation tasks, enabling the production of high-quality outputs that are both preference-aligned and test-time reliable.

    \vspace{0.5em}

    \item \textbf{\textit{Applying verifiers to both training and test time yields maximum enhancement.}}
    We observe the third combination approach delivers the most significant gains, outperforming the first approach by +8\%, and the second by +3\%. This suggests that, even with a model already aligned to preferences, the fine-tuned ORM can play complementary roles in the test-time decoding. These dual functions reinforce each other, highlighting the versatility of reward models in autoregressive image generation.
    
\end{itemize}

\section{Potential Assessment Reward Model}

From our comprehensive investigation, reward models prove valuable by enabling both decoding path selection and preference reward guidance. However, we still observe considerable room for enhancing reward models.

\paragraph{Limitation of ORM and PRM.}
\textit{1)} ORM showcases strong performance by selecting optimal final outputs, yet it lacks the capacity to provide fine-grained, step-wise evaluation at each generation step.
\textit{2)} While PRM has demonstrated effectiveness in understanding tasks such as mathematics, it is less suitable for autoregressive image generation. As analyzed in Sec.~\ref{s3.2}, PRM struggles with early-stage images that are too blurry for reliable evaluation, given that only a few regions are decoded. In later stages, images derived from similar previous steps lack sufficient distinction, challenging for PRM to discriminate.

\paragraph{PARM.}
Motivated by these observations, we propose the Potential Assessment Reward Model (PARM), a specialized reward model tailored for autoregressive image generation, as illustrated in Figure~\ref{rm_v1}. PARM combines the best of both worlds: \textit{1)} it operates adaptively in a step-wise manner, using a potential assessment mechanism to overcome PRM’s evaluation challenges; and \textit{2)} it performs a best-of-$N'$ selection across $N'$ ($N' \leq N$) high-potential reasoning paths, thus inheriting ORM’s advantage. Specifically, the methodology of PARM contains three progressive tasks:

\begin{enumerate}
    \item \textbf{Clarity Judgment.}
    In the best-of-$N$ setting, we first sample $N$ different reasoning paths for image generation.
    Then, at each intermediate step, PARM evaluates whether the partially generated image contains enough visual clarity to be meaningfully assessed, assigning a binary label If labeled `no', the model skips to the next step. If labeled `yes', the model proceeds to the next task for potential assessment.
    This pre-judgment prevents scoring on early, blurry images that lack informative content (as seen in PRM), ensuring only sufficiently clear steps are considered for rewarding.
    \vspace{0.5em}
    
    \item \textbf{Potential Assessment.}
    For each clear step that passes the clarity judgment, PARM assesses the potential of the current step to determine whether it can lead to a high-quality final image, again using a binary label. If labeled `no', the generation path is truncated immediately. If labeled `yes', the path is preserved to produce the final image. This approach is based on the observation that, once an image at a given step is clear enough to evaluate, its overall layout and structure are unlikely to change significantly in subsequent steps, making it a reliable candidate for potential assessment. This task helps identify promising intermediate steps, effectively pruning low-potential candidates during inference.
    
    \vspace{0.5em}
    \item \textbf{Best-of-$N'$ Selection.}
    After completing the above two tasks, suppose there are $N'$ high-potential paths remaining to produce the final images ($N' \leq N$).
    PARM then performs a best-of-$N'$ selection to identify the most promising image candidates as the output. If $N'=0$, the model defaults to selecting the reasoning path with the lowest probability of a `no' label as the output.
    This final task leverages ORM's global selection capabilities to ensure a high-quality generated image.
\end{enumerate}

\paragraph{PARM Ranking Data Curation.}

In Figure~\ref{early-later}, we illustrate why PRM is less suitable for autoregressive image generation. As shown, the early-stage images are too blurry for reliable evaluation, given that only a few regions are decoded, while the later-stage images derived from similar previous steps lack sufficient distinction, challenging for discrimination.
To integrate the advantage of both ORM and PRM, we propose Potential Assessment Reward Model (PARM) and curate a new ranking dataset with 400K instances by re-annotating the 13K text prompts from ORM ranking data. The dataset is structured into three subsets corresponding to the three tasks:

\begin{table*}[t]
  \centering
  \small
  % \vspace{-0.35cm}
  \caption{\textbf{Performance Comparison on the GenEval~\cite{ghosh2024geneval} Benchmark.} Compared to existing diffusion and autoregressive models, we investigate the potential of Chain-of-Thought (CoT) reasoning strategies in text-to-image generation. `Zs.', `Ft.', and `It. DPO' denote the zero-shot, fine-tuned verifiers, and iterative DPO~\cite{rafailov2024direct}, repsectively. \textbf{PARM} refers to our proposed Potential Assessment Reward Model specialized for autoregressive image generation. We adopt the best-of-$20$ selection for test-time verifiers, highlighting our best result in green and the previous best model in red.}
  % \label{table:all2}
  \vspace{0.1cm}
  \begin{adjustbox}{width=\textwidth}
  \begin{tabular}{l|ccc|cccccc|c}
    \toprule
    Model & \makecell*[c]{Test-time\\Verifier} & \makecell*[c]{Preference\\Alignment} & \makecell*[c]{Reward\\Guidance} &\makecell*[c]{Single\\object} & \makecell*[c]{Two\\object} & \makecell*[c]{Counting} & \makecell*[c]{Colors} & \makecell*[c]{Position} & \makecell*[c]{Attribute\\binding} & \textbf{Overall} \\
    \midrule
    PixArt-$\alpha$~\cite{chen2024pixartalpha} & - & - & - & 0.98 & 0.50 & 0.44 & 0.80 & 0.08 & 0.07 & 0.48 \\ 
    SD v2.1~\cite{rombach2022highresolutionimagesynthesislatent} & -& - & - & 0.98 & 0.51 & 0.44 & 0.85 & 0.07 & 0.17 & 0.50 \\ 
    DALL-E 2~\cite{ramesh2022hierarchicaltextconditionalimagegeneration} & -& - & - & 0.94 & 0.66 & 0.49 & 0.77 & 0.10 & 0.19 & 0.52 \\ 
    SDXL~\cite{podell2023sdxlimprovinglatentdiffusion} & - & - & -& 0.98 & 0.74 & 0.39 & 0.85 & 0.15 & 0.23 & 0.55 \\ 
    SD 3 (d=24)~\cite{esser2024scalingrectifiedflowtransformers} & - & - & -& 0.98 & 0.74 & 0.63 & 0.67 & 0.34 & 0.36 & \cellcolor{red!10}0.62 \\ 
    \midrule
    LlamaGen~\cite{sun2024beats} & - & - & -& 0.71 &0.34 &0.21 &0.58 &0.07 &0.04 &0.32 \\ 
    Chameleon~\cite{chameleonteam2024chameleonmixedmodalearlyfusionfoundation} & - & - & -& - & - & - & - & - & - & 0.39 \\ 
    LWM~\cite{liu2024worldmodelmillionlengthvideo} & - & - & -&  0.93 &0.41 &0.46 &0.79 &0.09 &0.15 &0.47 \\ 
    SEED-X~\cite{ge2024seedxmultimodalmodelsunified} & - & - & -&0.97 &0.58 &0.26 &0.80 &0.19 &0.14 &0.49\\
    \midrule
    \rowcolor{gray!12} \cellcolor{white}{\multirow{19}*{Show-o~\cite{xie2024show}}} &  - & - & -& 0.95 & 0.52 & 0.49 & 0.82 & 0.11 & 0.28 & 0.53 \\
    \cmidrule{2-11}
    & Zs. ORM &- &- & 0.99 & 0.63 & 0.63 & 0.84 & 0.19 & 0.39 & 0.61\\ 
    & Ft. ORM &- &- & 0.99 & 0.72 & 0.65 & 0.84 & 0.25 & 0.33 & 0.63 \\
    & Zs. PRM &- &- & 0.98&0.51&0.54&0.82&0.11&0.23&0.53\\ 
    & Ft. PRM &- &- & 0.98&0.55&0.54&0.83&0.13&0.29&0.55\\
    & \textbf{PARM} &-&-& 0.99 & 0.77 & 0.68 & 0.86 & 0.29 & 0.45 & 0.67 \\ 
    \cmidrule{2-11}
    & - &DPO &- & 0.96 & 0.70 & 0.50 & 0.82 & 0.30 & 0.43 & 0.62\\
    &-& It. DPO  &-& 0.98 & 0.72 & 0.53 & 0.84 & 0.40 & 0.46 & 0.65\\
    \cmidrule{2-11}
    & Zs. ORM & It. DPO & - & 0.99 & 0.79 &  0.63 & 0.85 & 0.44 & 0.50 & 0.70 \\ 
     & Ft. ORM& It. DPO & - & 0.98 & 0.80 & 0.62 & 0.83 & 0.59 & 0.54 & 0.72\\
      &\textbf{PARM}  &It. DPO &- &0.98 & 0.83 & 0.64 & 0.84 & 0.59 & 0.62 & 0.74 \\ 
      \cmidrule{2-11}

      & - &It. DPO&Ft. ORM& 0.98 & 0.80 & 0.62 & 0.83 & 0.59 & 0.54 &0.72\\ 
      & - &It. DPO&\textbf{PARM}& 0.97 & 0.75 & 0.60 & 0.83 & 0.54 & 0.53 & 0.69 \\ 
      \cmidrule{2-11}

     & Ft. ORM & It. DPO &Ft. ORM & 0.98 & 0.84 & 0.64 & 0.85 &  0.66 & 0.52 & 0.75\\ 
      &\textbf{PARM}&It. DPO &\textbf{PARM} & 0.99 & 0.86 & 0.67 & 0.84 & 0.66 & 0.64 & \cellcolor{green!15}0.77 \\ 
    \bottomrule 
  \end{tabular}
  \end{adjustbox}
  \label{tt3}
  % \vspace{-0.15cm}
  \vspace{0.3cm}
\end{table*}

\begin{itemize}
    \item \textbf{\textit{Clarity Judgment Data (120K):}} Through comprehensive analysis, we observe that the baseline model (Show-o) typically produces its first clear image between steps 8 and 12 within the 18-step generation, qualifying it for potential assessment. Based on this, we simplify the annotation by labeling steps after 11 as `yes' and those before 10 as `no'. Although this approach is static, the trained PARM acquires generalization skills to adaptively identify the first `yes' label within steps 8$\sim$12 during inference. The data format is shown below:

\begin{formal}
    \textbf{Instruction:} \textit{``\textbf{ \{image\}} This image is a certain step in the text-to-image generation process with a prompt: \textbf{\{prompt\}}. It is not the final generated one, and will keep iterating better.               Do you think this image can be used to judge whether it has the potential to iterate to the image satisfied the prompt? (The image, which needn't to be confused  but can be clear and basically judged the object, can be used to judge the potential) Answer yes or no without explanation.''}\\
    \\
    \textbf{Response:} \textit{``Yes''} or \textit{``No''}
\end{formal}

\item \textbf{\textit{Potential Assessment Data (80K): }} We assign intermediate images from steps after 11 with a `yes' or `no' label, which is based on the final output label of that path in the ORM data annotation. In practice, if the previous clarity judgment task yields `yes', the data of this task is organized as a follow-up question-answering within a multi-turn conversation. The data sample of this task is formulated as:
\begin{formal}
    \textbf{Instruction:} \textit{``\textbf{ \{image\}} Do you think whether the image has the potential to iterate to the image satisfied the prompt? Please answer yes or no without explanation.''}\\
    \\
    \textbf{Response:} \textit{``Yes''} or \textit{``No''}
\end{formal}

\item \textbf{\textit{Best-of-$N'$ Selection Data (200K):}} We directly utilize the labels in the ORM ranking dataset, with the format as
\begin{formal}
    \textbf{Instruction:} \textit{``\textbf{ \{image\}} This image is generated by a prompt: \textbf{\{prompt\}}. Does this image accurately represent the prompt? Please answer yes or no without explanation.''}\\
    \\
    \textbf{Response:} \textit{``Yes''} or \textit{``No''}
\end{formal}

\end{itemize}
%%%%%%%%%%%%%%%%%%%%%%%%%%%

% \vspace{-0.3cm}
\paragraph{Experiments and Insights.}
With the new reward model, we revisit our previous investigation by applying PARM to different approaches enhancing autoregressive image generation. The observations are summarized below:

\begin{table*}[t]
  \centering
  \small
  \caption{
  \textbf{Generalization Results of DPO, Finetuned ORM, and PARM on Sequential Autoregressive Generative Models, LlamaGen~\cite{sun2024autoregressive} and Janus-Pro~\cite{chen2025janus}.} `Zs.' and `Ft.' denote the zero-shot, fine-tuned verifiers. \textbf{PARM} refers to our proposed Potential Assessment Reward Model specialized for autoregressive image generation. We adopt the best-of-$20$ selection for test-time verifiers. Our methods consistently improve performance across both models, highlighting their model-agnostic and plug-and-play nature.}
  \vspace{0.1cm}
  \begin{adjustbox}{width=0.9\textwidth}
  \begin{tabular}{l|c|cccccc|c}
    \toprule
    Model & Method &\makecell*[c]{Single\\object} & \makecell*[c]{Two\\object} & \makecell*[c]{Counting} & \makecell*[c]{Colors} & \makecell*[c]{Position} & \makecell*[c]{Attribute\\binding} & \textbf{Overall} \\
    \midrule
    \rowcolor{gray!12} \cellcolor{white}{\multirow{4}*{LlamaGen~\cite{sun2024autoregressive}}} & - &0.71 &0.34 &0.21 &0.58 &0.07 &0.04 &0.32 \\
    \cmidrule{3-9}
    & Zs. ORM & 0.75 & 0.37 & 0.24 &0.61 &0.08 & 0.06 &0.35 \\ 
    & Ft. ORM & 0.80 &0.41 &0.29 &0.65 &0.10 &0.08 &0.39\\
    & DPO  &0.78 &0.43 &0.30 &0.63 &0.19 &0.14 &0.41 \\
    & \textbf{PARM} &0.84 &0.45 &0.31 &0.67 &0.20 &0.21 &\cellcolor{green!15}0.46 \\ 
    \midrule
    \rowcolor{gray!12} \cellcolor{white}{\multirow{4}*{Janus-Pro~\cite{chen2025janus}}} &- &0.99	&0.89	&0.59	&0.90	&0.79	&0.66	&0.80\\
    \cmidrule{3-9}
    & Zs. ORM &0.99	&0.91	&0.77	&0.90	&0.79	&0.67	&0.84 \\ 
    & Ft. ORM &0.99	&0.93	&0.82	&0.92	&0.91	&0.76 &0.89 \\
    &DPO &0.99	&0.89	&0.65	&0.92	&0.82	&0.72	&0.83 \\
    &\textbf{PARM} &1.00	&0.95	&0.80	&0.93	&0.91	&0.85	&\cellcolor{green!15}0.91 \\ 
    \bottomrule 
  \end{tabular}
  \end{adjustbox}
  \label{tt5}
  % \vspace{-0.15cm}
  \vspace{0.3cm}
\end{table*}

\begin{figure}[t]
  \centering
  \includegraphics[width=0.95\linewidth]{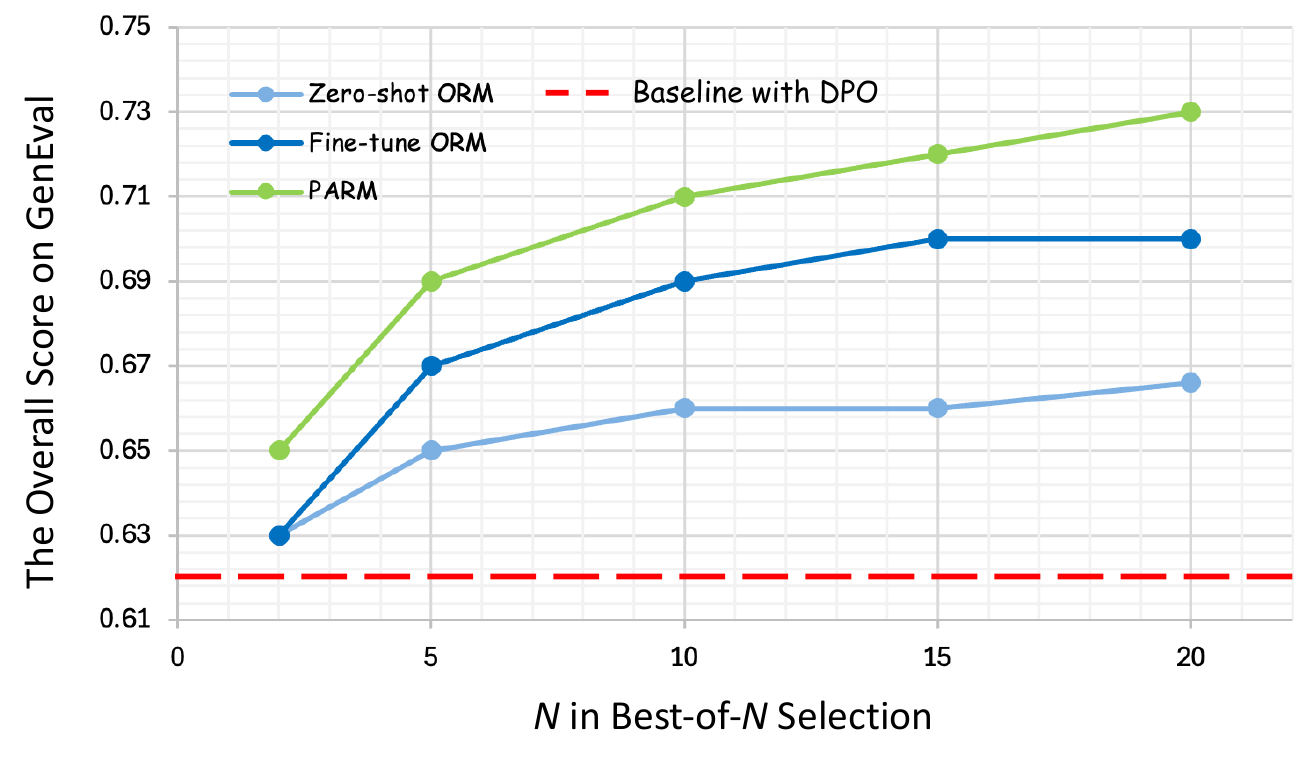}
  \vspace{0.2cm}
  \caption{\textbf{Comparison of Reward Models as Test-time Verifiers with DPO Alignment.} We adopt Show-o~\cite{xie2024show} with DPO alignment as the `Baseline with DPO' and evaluate Best-of-$N$ selection on the GenEval~\cite{ghosh2024geneval} benchmark.}
  \label{best1}
\end{figure}

\begin{figure*}[t]
  \vspace{0.3cm}
  \centering
  \includegraphics[width=\linewidth]{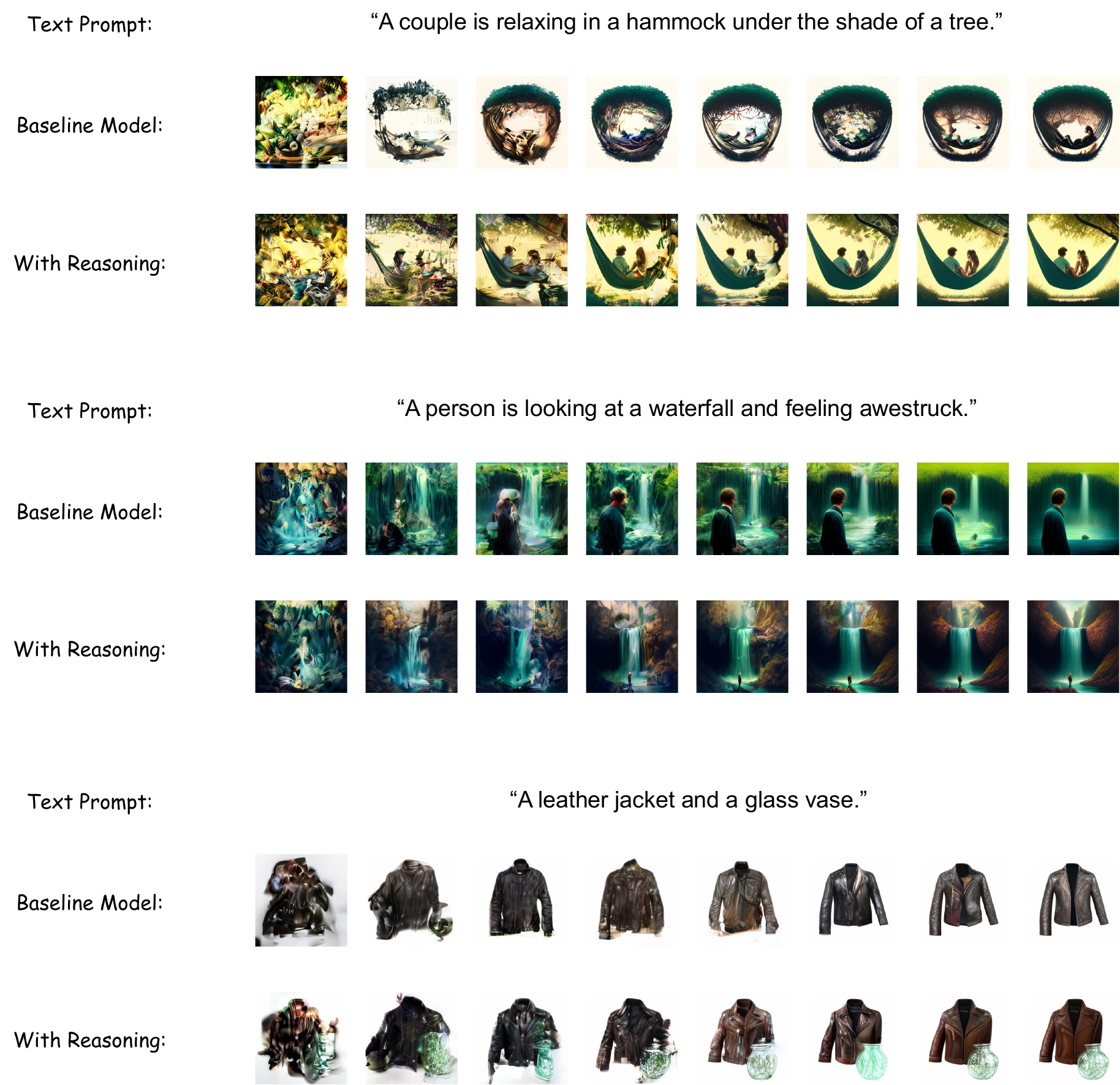}
  \vspace{0.3cm}
  \caption{\textbf{Qualitative Results using Our Reasoning Strategies.} Show-o~\cite{xie2024show} is adopted as the baseline model, and compared to our best-performing reasoning strategy: integrating PARM with iterative DPO for both reward model guidance and test-time verification.}
  \label{sup1}
\end{figure*}

\begin{figure*}[t]
  \vspace{0.3cm}
  \centering
  \includegraphics[width=\linewidth]{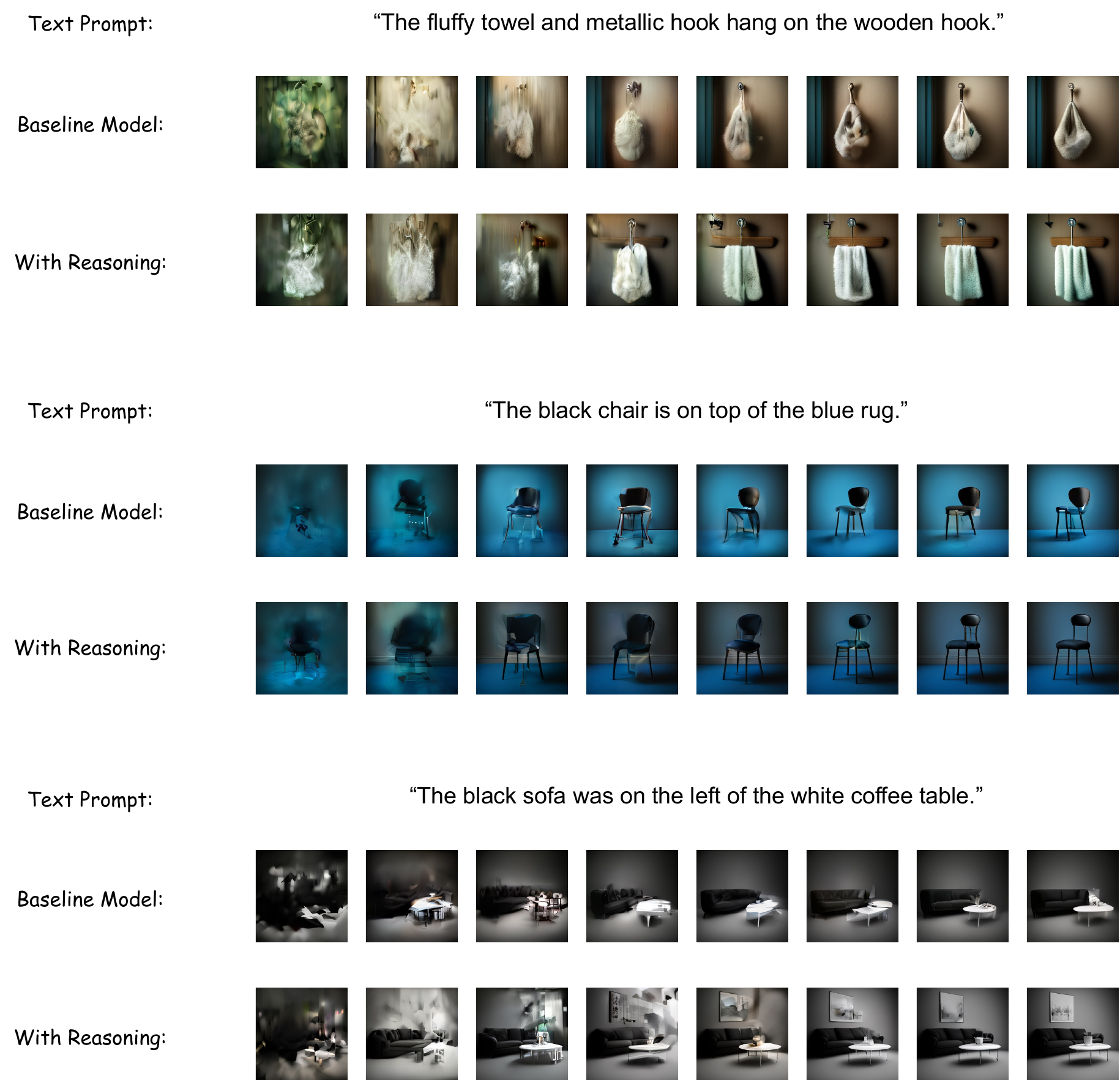}
  \vspace{0.3cm}
  \caption{\textbf{Qualitative Results using Our Reasoning Strategies.} Show-o~\cite{xie2024show} is adopted as the baseline model, and compared to our best-performing reasoning strategy: integrating PARM with iterative DPO for both reward model guidance and test-time verification.}
  \label{sup2}
\end{figure*}

\begin{figure*}[t]
  \vspace{0.3cm}
  \centering
  \includegraphics[width=\linewidth]{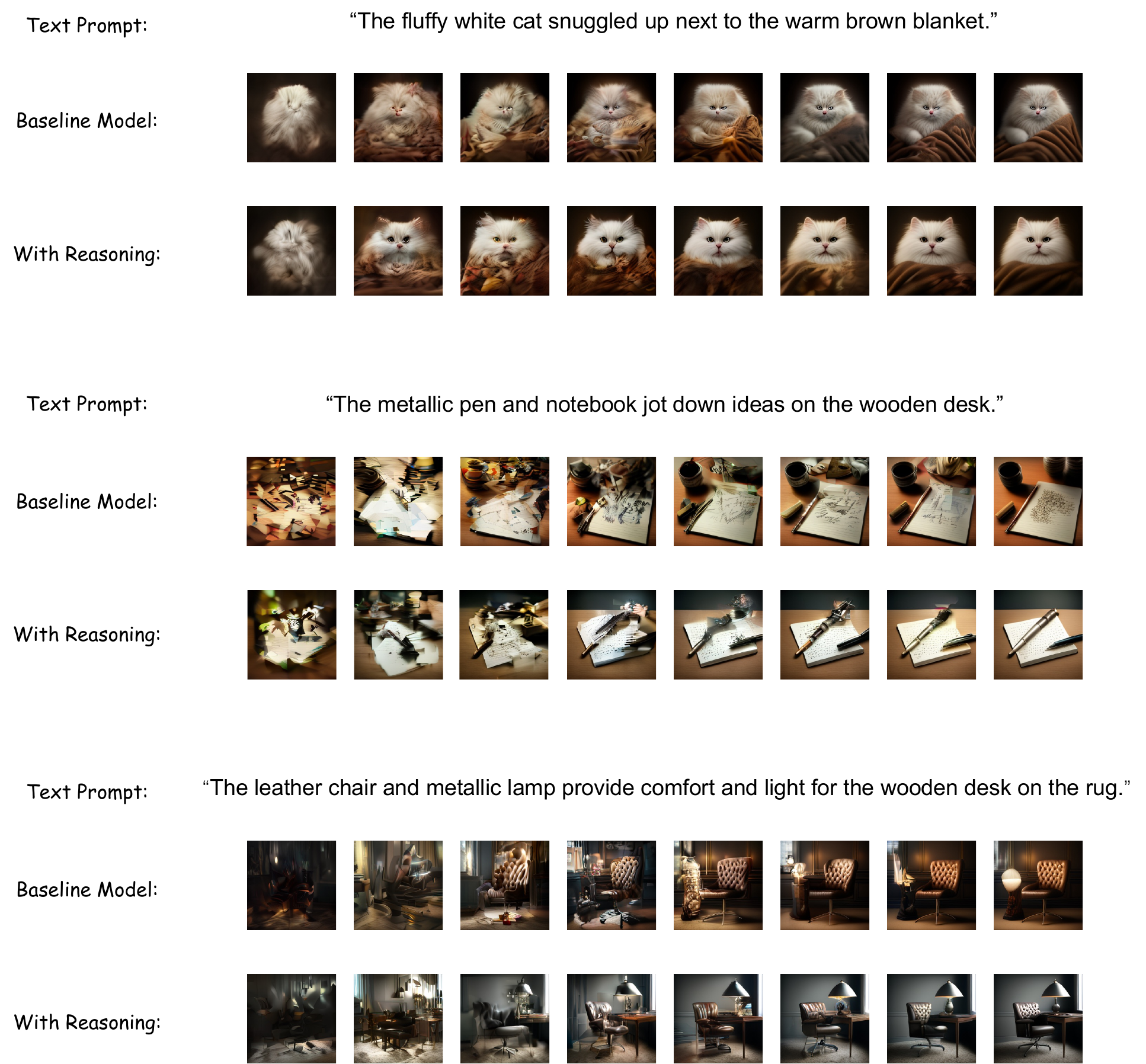}
  \vspace{0.3cm}
  \caption{\textbf{Qualitative Results using Our Reasoning Strategies.} Show-o~\cite{xie2024show} is adopted as the baseline model, and compared to our best-performing reasoning strategy: integrating PARM with iterative DPO for both reward model guidance and test-time verification.}
  \label{sup3}
\end{figure*}

\begin{figure*}[t]
  \vspace{0.3cm}
  \centering
  \includegraphics[width=\linewidth]{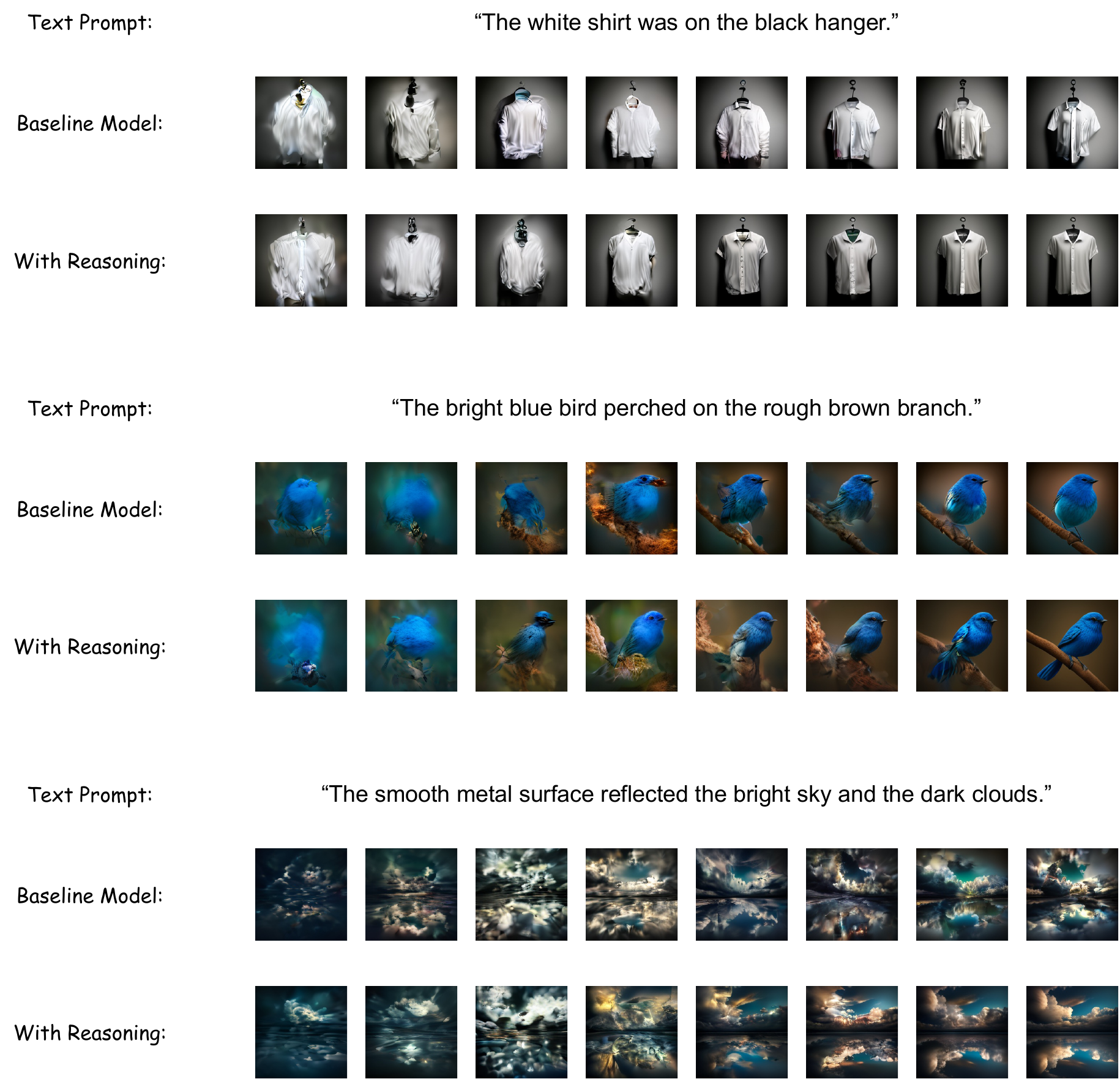}
  \vspace{0.3cm}
  \caption{\textbf{Qualitative Results using Our Reasoning Strategies.} Show-o~\cite{xie2024show} is adopted as the baseline model, and compared to our best-performing reasoning strategy: integrating PARM with iterative DPO for both reward model guidance and test-time verification.}
  \label{sup4}
\end{figure*}

\begin{figure*}[t]
  \vspace{0.3cm}
  \centering
  \includegraphics[width=\linewidth]{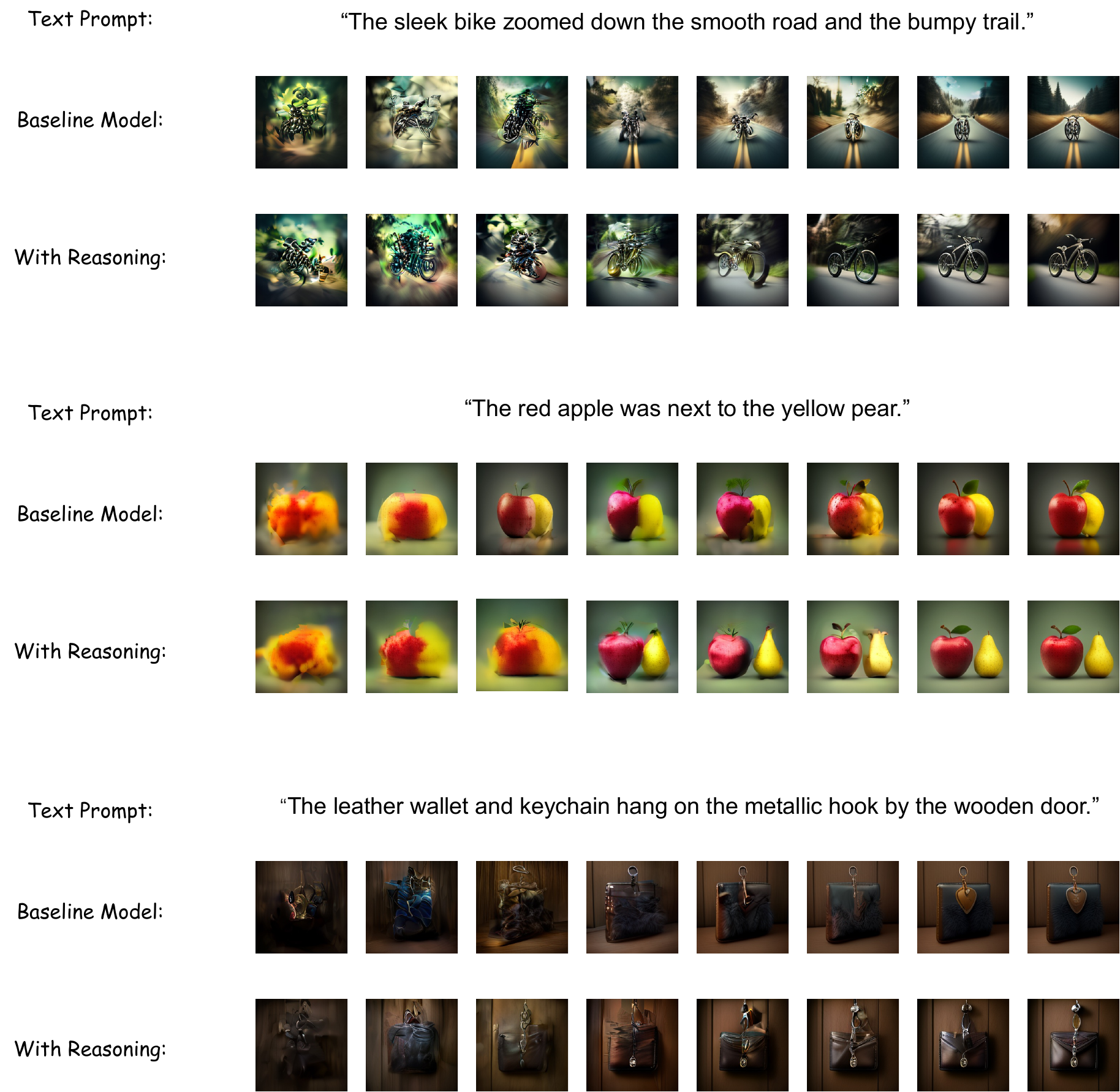}
  \vspace{0.3cm}
  \caption{\textbf{Qualitative Results using Our Reasoning Strategies.} Show-o~\cite{xie2024show} is adopted as the baseline model, and compared to our best-performing reasoning strategy: integrating PARM with iterative DPO for both reward model guidance and test-time verification.}
  \label{sup5}
\end{figure*}

\begin{itemize}
    \item \textbf{\textit{PARM demonstrates the best-performing reward model across different strategies.}}
    Table~\ref{tt3} and Figure~\ref{bon} present the effectiveness of PARM as test-time verifiers, significantly outperforming other reward models, e.g., +6\% to the fine-tuned ORM.
    Additionally, PARM scales effectively with increasing $N$, highlighting its potential for further improvement with larger test-time computation.
    PARM also outperforms iterative DPO, the enhanced preference alignment strategy with refined data. Furthermore, PARM better harnesses the complementary strengths with post-training, consistently attaining higher integration scores than the fine-tuned ORM. These results underscore PARM's capability as a versatile, robust reward model for autoregressive image generation.
    
    \vspace{0.5em}
    
    \item \textbf{\textit{With PARM, our baseline model (Show-o) is enhanced to achieve leading generation performance.}}
    Compared to other image generation models in Table~\ref{tt3}, our best-performing configuration, i.e., integrating PARM with iterative DPO in both post-training and test-time, achieves a score of 77\%, improving the baseline by +24\% and surpassing the advanced Stable Diffusion 3~\cite{esser2024scalingrectifiedflowtransformers} by +15\%. 
    In particular, substantial gains are observed in `Two Obj.', `Colors', `Position', and `Attribute binding' emphasizing the robustness in handling challenging compositional generation. In Figures~\ref{best1} and \ref{best2}, we present the performance of test-time verification integrated with DPO~\cite{rafailov2024direct} and iterative DPO, respectively, instead of the test-time verification only in Figure~\ref{intro_vis}. As shown, our propose PARM both achieves the best results as the $N$ increases for best-of-$N$ selection. Additionally, in Figures~\ref{sup1}, \ref{sup2}, \ref{sup3}, \ref{sup4}, and \ref{sup5}, we showcase extensive qualitative examples. We observe that the baseline model often generates inaccurate spatial relationships, strange appearances, and object attributes. In contrast, our approach consistently mitigates such issues, ensuring that the spatial relations, object features, and overall fidelity to the text prompt are preserved. 

\begin{figure}[t]
  \centering
  \includegraphics[width=0.95\linewidth]{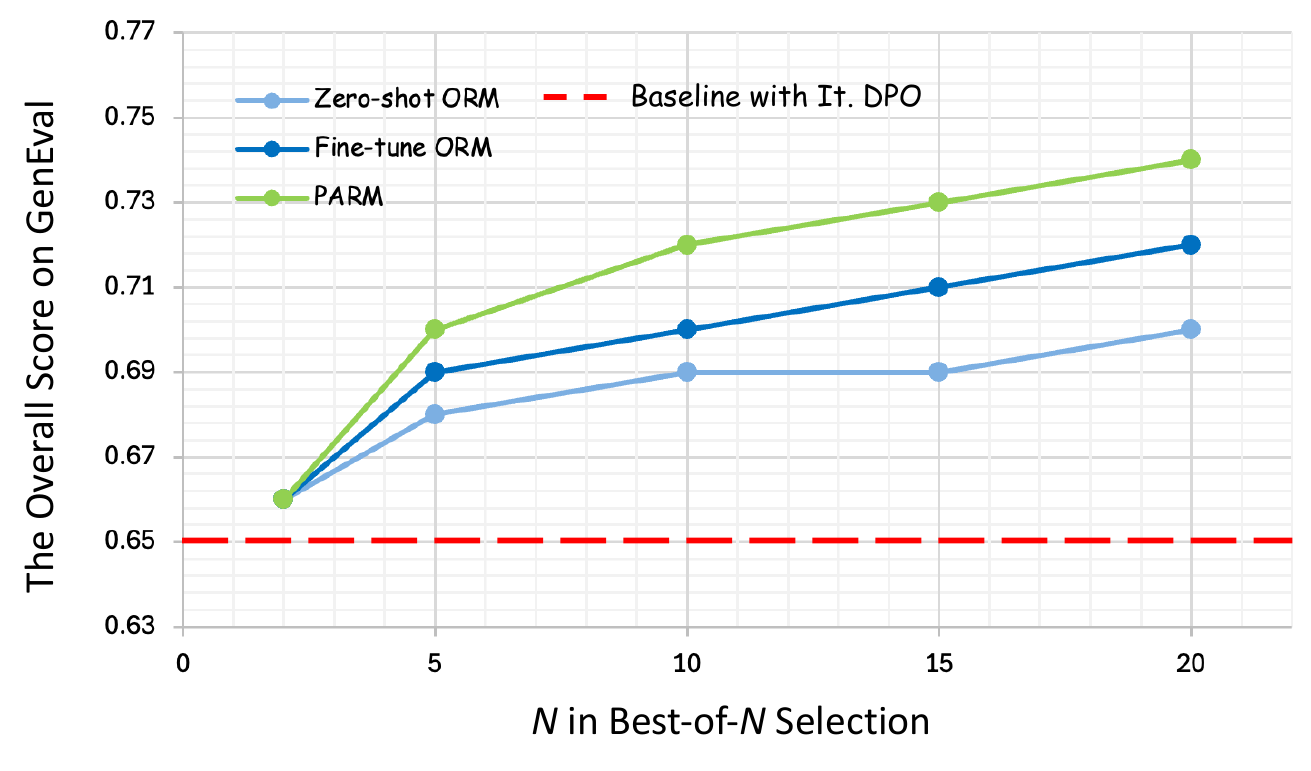}
  \vspace{0.2cm}
  \caption{\textbf{Comparison of Reward Models as Test-time Verifiers with Iterative DPO Alignment.} We adopt Show-o~\cite{xie2024show} with iterative DPO alignment as the `Baseline with It. DPO' and evaluate Best-of-$N$ selection on GenEval~\cite{ghosh2024geneval}.}
  \label{best2}
\end{figure}

    \vspace{0.5em}
    \item \textbf{\textit{Generalization to Sequential Autoregressive Generative Models.}}
While our method is validated on Show-o, which uses adaptive generation ordering with a discrete diffusion approach, we further extend it to sequential autoregressive models that generate image tokens via standard next-token prediction. Specifically, we apply DPO, Finetuned ORM, and PARM to LlamaGen-3B~\cite{sun2024autoregressive} and Janus-Pro-7B~\cite{chen2025janus}, both of which follow token-by-token generation in a fixed order. As shown in Table~\ref{tt5}, our method consistently improves performance across both models without requiring retraining or architectural changes, demonstrating its generality and plug-and-play nature.

\end{itemize}

\section{Potential Assessment Reward Model ++}

Besides sequential step-by-step reasoning, humans often engage in a reflection process to verify the correctness of their previous thoughts. To explore its potential in image generation, we introduce PARM++, as shown in Figure~\ref{rm_v1}, which enhances PARM with a reflection mechanism to refine the text-to-image quality.

\paragraph{Reflection in Image Generation.}
The reflection strategy has also been recently applied in LLMs~\cite{kumar2024training,huang2023large}, resulting in improved performance through self-correction. Unlike LLMs, which can produce and interpret free-form language to review and refine their outputs, image generation models typically rely on a text prompt (often descriptive rather than instructive) as input and output only the image modality. Consequently, the reflection capability must be primarily handled by an external reward model, tasked with identifying misalignments and providing explanations. Additionally, the image generation model itself also needs to be fine-tuned to effectively understand and respond to these reflection texts for self-correction.

\paragraph{PARM++.}
After selecting the optimal output from $N'$ image candidates, PARM++ incorporates a reflection evaluation task that examines the alignment between the final image and the input text prompt.
If the image satisfies the alignment criteria, PARM++ outputs a `yes' and considers it the final result. Otherwise, it provides a detailed analysis of the discrepancies, including specific reasons for the misalignment between the image and the prompt. 
Then, we feed three inputs into the image generation model to self-correct its image output, including the original text prompt, the previously generated suboptimal image, and the identified misalignment reasons. This iterative refinement process continues until PARM++ produces a `yes' in the reflection evaluation, thereby progressively improving both the visual fidelity and the image-text correspondence. We set the maximum number of reflection iterations to 3.

\begin{figure}[t]
\hspace{0.65cm}
  \includegraphics[width=0.85\linewidth]{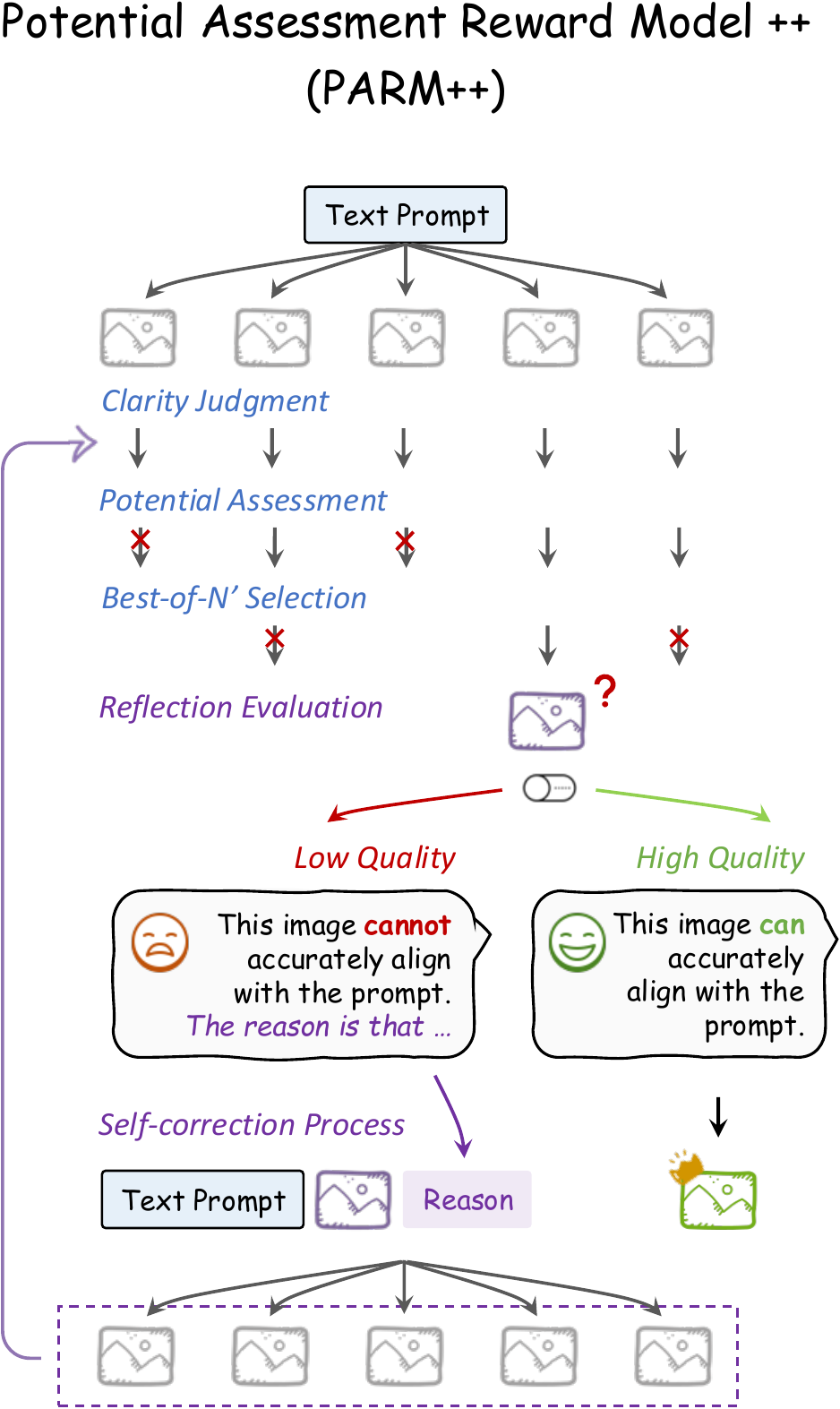}
  \vspace{0.3cm}
  \caption{\textbf{The Reflection Mechanism in Potential Assessment Reward Model ++ (PARM++).} As an upgraded version of PARM, PARM++ incorporates a reflection evaluation task, enabling the generative model to self-correct its low-quality images.}
  \label{parm++}
\end{figure}

\paragraph{PARM++ Ranking Data Curation.}
Building on the 400K dataset used for PARM, we curate an additional 120K instances for the reflection evaluation task, resulting in a total of 520K data points for training PARM++. For the negative data, we select samples labeled as `no' from the ORM ranking dataset, representing low-quality images to be refined, and leverage GPT-4o~\cite{openai2024gpt4o} to provide concise annotations detailing the image-text discrepancies. For the positive data, we directly utilize samples labeled as `yes' from the ORM ranking dataset, representing high-quality images passing the reflection evaluation. The ratio of negative to positive instances is approximately $80\%: 20\%$ in the dataset.

\begin{figure*}[t]
  \includegraphics[width=\linewidth]{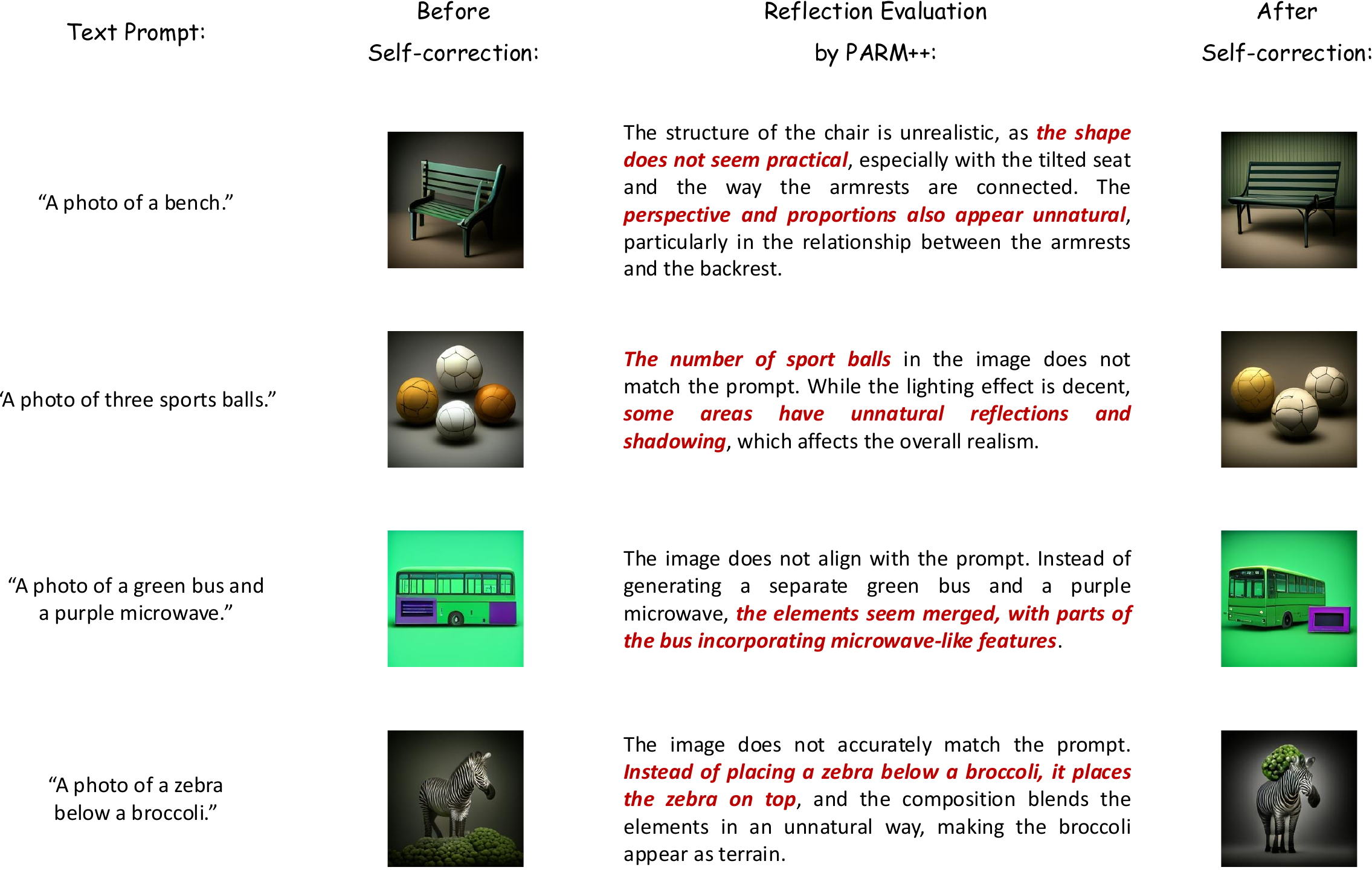}
  \vspace{0.1cm}
  \caption{\textbf{Qualitative Results with Reflection in PARM++.} The proposed PARM++ incorporates a reflection evaluation stage to detect text-image misalignments and provides detailed explanations to guide the self-correction process in autoregressive image generation models.}
  \vspace{-0.3cm}
  \label{parm++_vis}
\end{figure*}

\begin{table*}[t]
  \centering
  \small
  % \vspace{0.3cm}
  \caption{\textbf{PARM++ with Reflection as Test-time Verifiers.} We evaluate text-to-image generation on the GenEval~\cite{ghosh2024geneval} benchmark and adopt Show-o~\cite{xie2024show} as the autoregressive baseline model. `Ft. ORM' and `PARM' denote the fine-tuned ORM and our Potential Assessment Reward Model, respectively. We adopt the best-of-$N$ selection, setting $N=20$, and highlight the best-performing variant in green.}
  \vspace{0.1cm}
  \begin{adjustbox}{width=\textwidth}
  \begin{tabular}{l|cc|cccccc|c}
    \toprule
    \makecell*[l]{Reasoning\\Strategy} &\makecell*[c]{Test-time\\Verifier}& \makecell*[c]{Reflection}&\makecell*[c]{Single\\object} & \makecell*[c]{Two\\object} & \makecell*[c]{Counting} & \makecell*[c]{Colors} & \makecell*[c]{Position} & \makecell*[c]{Attribute\\binding} & \textbf{Overall} \\
    \midrule
    \rowcolor{gray!12}Baseline& - &- & 0.95 & 0.52 & 0.49 & 0.82 & 0.11 & 0.28 & 0.53 \\
    \cmidrule{1-10}
    \multirow{6}*{\makecell*[l]{Self-correction\\Fine-tuning}}& - &- & 0.92 & 0.50 & 0.47 & 0.79 & 0.10 & 0.27 & 0.51 \\
    \cmidrule{2-10}
    & Ft. ORM &- & 0.97 & 0.57 & 0.54 & 0.86 & 0.20 & 0.35 & 0.58 \\
    & PARM &- & 0.98 & 0.65 & 0.61 & 0.91 & 0.26 & 0.40 & 0.64 \\
    \cmidrule{2-10}
    & \bf PARM++ &- & 0.97 & 0.60 & 0.58 & 0.89 & 0.24 & 0.38 & 0.61 \\
    & \bf PARM++ &\checkmark & 0.99 & 0.71 & 0.69 & 0.95 & 0.36 & 0.49 &\cellcolor{green!15}0.70 \\
    \bottomrule
  \end{tabular}
  \end{adjustbox}
  \label{parm+++}
  \vspace{0.3cm}
\end{table*}

\paragraph{Self-correction Fine-tuning.}
As the baseline model, e.g., Show-o, is not pre-trained to refine low-quality images based on textual instructions, we specifically fine-tune Show-o to endow it with the capability of self-correcting generated images. Fortunately, Show-o supports simultaneous inputs of both text and images, enabling image refinement guided by textual feedback. To curate the training data, we extract 10K instance groups from the PARM++ ranking dataset. Each group consists of a text prompt, a low-quality (negative) image, a high-quality (positive) image, and the annotated reflection reasons. This dataset is used to fine-tune Show-o for iterative image refinement, progressively improving the quality and alignment of the generated images.

\paragraph{Experiments and Insights.}
In Table~\ref{parm+++}, we showcase the results of reflection and self-correction using PARM++.
Note that we do not incorporate any DPO alignment upon Show-o, in case of the training conflict with self-correction fine-tuning. The observations are summarized below:

\begin{itemize}
    \item \textbf{\textit{The reflection mechanism in PARM++ significantly enhances image quality.}} As an ablation, when no reflection is performed (`-' in the table), PARM++ performs slightly worse than PARM due to the integration of reflection-specific training data. However, when the reflection mechanism is enabled (`$\checkmark$' in the table), the overall score on GenEval improves substantially by +10\%, underscoring the effectiveness of the reflection strategy in PARM++. In Figure~\ref{parm++_vis}, we illustrate a comparison of images before and after the self-correction process. Guided by the reflection evaluation of PARM++, the self-corrected images address issues such as unrealistic elements, incorrect numbers, wrong colors, and improper object layouts.

    \vspace{0.5em}

    \item \textbf{\textit{The Self-correction Fine-tuning slightly impacts the baseline model's performance.}}
    While PARM++ ultimately improves performance through its reflection mechanism, the self-correction fine-tuning introduces a minor trade-off, resulting in a -2\% drop in accuracy on GenEval. This decline occurs because training for new capabilities inevitably influences the model's original knowledge. We anticipate that this limitation can be mitigated in the future with the advancement of more general large models, which are expected to inherently possess the robustness to handle multi-modal information seamlessly.
\end{itemize}

\section{Conclusion}
In this work, we investigate the adaption and potential of CoT reasoning strategies in autoregressive image generation. Through a systematic investigation, we demonstrate that different reasoning strategies can effectively improve image generation, e.g., test-time verification, preference alignment, and their integration.
Given our observation, we further introduce
two tailored reward models for autoregressive image generation, termed Potential Assessment Reward Model (PARM) and PARM++, which evaluate the step-wise generation for adaptive reward scoring, and incorporate a reflection mechanism for self-corrected image generation. Our experiments underscore the promise of CoT reasoning in autoregressive image generation, advancing this field in new directions.

\clearpage

\clearpage

\clearpage

\clearpage

\clearpage
\bibliography{sn-bibliography}
\end{document}